\documentclass[final]{cvpr}

\usepackage{times}
\usepackage{epsfig}
\usepackage{graphicx}
\usepackage{amsmath}
\usepackage{amssymb}

\usepackage{amsthm}
\usepackage{multirow}
\usepackage{multicol}
\usepackage{tabularx}
\usepackage{booktabs}
\usepackage{enumitem}
\usepackage{array}
\usepackage{color, colortbl}
\usepackage{fixltx2e}
\usepackage{caption}

\usepackage[ruled,vlined,linesnumbered]{algorithm2e}
\usepackage{subcaption}
\makeatletter
\@namedef{ver@everyshi.sty}{}
\makeatother
\usepackage{pgffor}
\usepackage{caption}

\usepackage{ dsfont }
\usepackage{cuted}
\usepackage[page]{appendix}
\usepackage[pagebackref=true,breaklinks=true,colorlinks,bookmarks=false]{hyperref}

\def\cvprPaperID{7375} % *** Enter the CVPR Paper ID here
\def\confYear{CVPR 2021}

\graphicspath{{figures_arxiv/}}
\usepackage{xstring}
\newcommand{\mymacro}[1]{%
    \StrCut[1]{#1}{-}\csA\csB 
    \StrCut[1]{\csB}{-}\csC\csD
    \StrCut[1]{\csD}{-}\csE\csF
    \StrCut[1]{\csF}{-}\csG\csH
    Org.: \csC \hphantom|\hphantom| UV: \csE \hphantom|\hphantom| Frontalized: \csG
}

\DeclareMathOperator*{\argmin}{arg\,min}

\begin{document}

%%%%%%%%% TITLE
\title{OSTeC: One-Shot Texture Completion}

\author{Baris Gecer, ~~Jiankang Deng, ~~and Stefanos Zafeiriou\\
Imperial College London\\
{\tt\small \{b.gecer, j.deng16, s.zafeiriou\}@imperial.ac.uk}
}

\maketitle
%%%%%%%%%%%%%%%%%%%%%%% FIRST PAGE FIGURE %%%%%%%%%%%%%%%%%%%%%%%%%
%-----------------------------------------------------------------------------------------%
\begin{strip}\centering
\vspace{-1cm}
\includegraphics[width=1.0\textwidth]{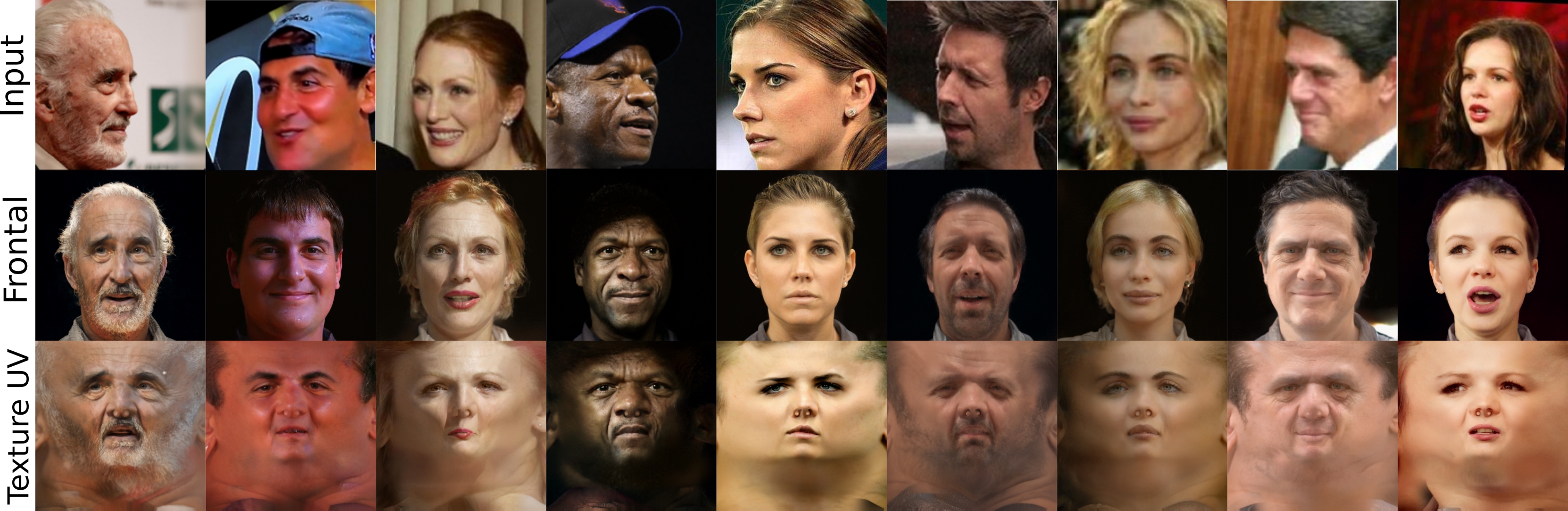}
\captionof{figure}{Face frontalization and UV texture completion by our approach. The first row is the input, the second row is the frontalization result, and the third row is the completed UV texture. The proposed method can produce photo-realistic and identity-preserved full UV textures even under extreme poses.}
\label{fig:teaser}
\end{strip}
%-----------------------------------------------------------------------------------------%

%%%%%%%%% ABSTRACT
\begin{abstract}
The last few years have witnessed the great success of non-linear generative models in synthesizing high-quality photorealistic face images. Many recent 3D facial texture reconstruction and pose manipulation from a single image approaches still rely on large and clean face datasets to train image-to-image Generative Adversarial Networks (GANs). Yet the collection of such a large scale high-resolution 3D texture dataset is still very costly and difficult to maintain age/ethnicity balance. Moreover, regression-based approaches suffer from generalization to the in-the-wild conditions and are unable to fine-tune to a target-image. In this work, we propose an unsupervised approach for one-shot 3D facial texture completion that does not require large-scale texture datasets, but rather harnesses the knowledge stored in 2D face generators. The proposed approach rotates an input image in 3D and fill-in the unseen regions by reconstructing the rotated image  in  a  2D  face  generator, based  on  the visible parts. Finally, we stitch the most visible textures at different angles in the UV image-plane. Further, we frontalize the target image by projecting the completed texture into the generator. The qualitative and quantitative experiments demonstrate that the completed UV textures and frontalized images are of high quality, resembles the original identity, can be used to train a texture GAN model for 3DMM fitting and improve pose-invariant face recognition.\footnote{Project Page: \url{https://github.com/barisgecer/OSTeC}}

\end{abstract}

%%%%%%%%% BODY TEXT
\section{Introduction}
The problem of 3D face texture completion (as shown in Fig.~\ref{fig:3DMM}) refers generally to the problem of recovering near ear-to-ear visible and non-visible colour from a single image \cite{deng2018uv} in a “canonical”, deformation-free parameterization of the face surface (usually referred as UV-space). A very similar problem is that of producing arbitrary face rotations from a single image \cite{zhou2020rotate,blanz2003face}. Both of the above problems have important applications in many different domains of face analysis such as pose-invariant face recognition \cite{deng2018uv,blanz2003face}, as well developing of 3D Morphable Model (3DMM) algorithms \cite{booth20173d,gecer2019ganfit} and creating complete head avatars from single images \cite{lattas2020avatarme}. That is why 3D face texture completion, as well as, producing face rotations has been very popular in the intersection of machine learning and computer vision, offering an important application domain to the advancements of machine learning in each era (from robust component analysis \cite{sagonas2017robust} to modern deep learning \cite{deng2018uv,zhou2020rotate}).

The problem of predicting the missing colour in the texture coordinated of the UV space or predicting a new view from a single image has been the application domain of many machine learning algorithms starting from simple nearest-neighbour interpolation, (\ie Fig.~\ref{fig:3DMM_uv}), regression techniques using linear-statistical priors (e.g., Robust Principal Component Analysis \cite{candes2011robust}) to modern deep learning regression techniques such as image-to-image translation models using conditional Generative Adversarial Networks (GANs)~\cite{isola2017image}. The problem has been modeled as fully supervised, \ie the regression model was trained with pairs of missing and complete 3D facial texture \cite{deng2018uv}, or recently using self-supervised methods and image rendering \cite{zhou2020rotate}. Nevertheless, fully-supervised or self-supervised, to the best of our knowledge, all current methods belong in the family of regression techniques. 

Contrary to the above, we take a radically different line of work in this paper: We propose to re-think the 3D facial texture prediction and rotation generation as an optimisation problem and design our method as a one-shot texture completion approach. One of the key problems of regression-based approaches such as \cite{zhou2020rotate} is that they may lose the identity because the function they learn is quite generic. Contrary, our approach optimises, along-side many other functions, identity-related features. Our method produces visually stunning results in both 3D texture completion as well as frontalization (for some results please inspect Fig.~\ref{fig:teaser}). Another by-product of our method is a 3D texture model learned from in-the-wild images that, as we show, can be used for training state-of-the-art 3D face reconstruction algorithms such as GANFit~\cite{gecer2019ganfit} (which was trained with around 10K 3D faces captured in well-controlled conditions which are not released to the public). 

In short, the contributions of our paper are as follows:
\begin{itemize}
    \item We re-design the problem of 3D facial texture completion as a one-shot optimisation-based approach. We propose a well-engineered novel methodology and cost function suitable for the task. 
    \item We capitalize on the power of 2D face generators to recover unseen part of 2D face by rotating it in 3D. So that, there would no need for 3D data collection.
    \item We show the effectiveness of the proposed approach in qualitative and quantitative experiments. Additionally, we apply the method to many in-the-wild images in order to train a large-scale prior of the 3D facial texture which we use to train state-of-the-art 3D face reconstruction algorithms.
\end{itemize}

\section{Related Work}
\noindent{\bf Face Generation, Manipulation \& Rotation :}
In just a few years, the quality of face generations by GANs have improved incredibly~\cite{karras2017progressive,karras2018style,karras2020analyzing}. The recently proposed StyleGANv2~\cite{karras2020analyzing} has shown high-quality 2D face generations up to $1024 \times 1024$ by eliminating artefacts that appear in the previous results. Many follow up works \cite{shen2019interpreting,tewari2020stylerig,abdal2019image2stylegan,abdal2019image2stylegana,gu2020image,gu2020image,qian2019make} could successfully project real images over its latent space and perform semantic manipulation. This indicates that one can utilize StyleGAN generator as a 2D facial texture prior. In this study, we exploit this finding for image inpainting to recover the unseen part of a 2D face.

One of the commonly manipulated facial attributes is the pose, especially to a frontal view for its applications in face recognition and normalization. Unfortunately, above mentioned latent space manipulation methods are either struggling to disentangle other attributes from the latent parameters or having difficulty to project an in-the-wild image to this space. Even if it is possible to achieve excellent reconstruction by projecting to the extended latent space ($\mathbb{R}^{18 \times 512}$) of StyleGAN~\cite{abdal2019image2stylegan,abdal2019image2stylegana}, this enforcement exhaust its semantic meaning, therefore, become non-functional for frontalization. In fact, one can project a cat image to a StyleGAN trained on human faces by these approaches.

A large body of work addresses this problem by image-to-image translation GANs~\cite{tran2017disentangled,yin2017towards,huang2017beyond,hu2018pose,tran2018representation,qian2019unsupervised}. Many of these approaches utilize paired datasets in a supervised setting which does not generalize well to in-the-wild settings. A recent work~\cite{zhou2020rotate} proposed a self-supervised training approach which perturbs images by 3D rotation to generate training pairs automatically. Nevertheless, these regression-based methods suffer from generalization and fall behind the optimization-based approaches which can fine-tune for any target image.

\noindent{\bf 3D Texture Completion :}
Modelling and synthesis of faces have been extensively studied in 3D as well~\cite{blanz1999morphable,egger20203d,gecer2018facegan,garbin2020high,gecer2020synthesizing,shamai2019synthesizing, gecer2019ganfit}. Nevertheless, generations from these models have been far from being photorealistic. Therefore, there have been some works that proposed to complete a partially visible appearance of 2D images to a 3D appearance maps~\cite{li2017generative,deng2018uv}. The most recent one~\cite{deng2018uv} trains an image-to-image translation network supervised by a set of controlled datasets, failing to generate high-quality images for in-the-wild settings. Although the proposed approach tackles the problem of texture completion, it brings a new perspective which is formulating texture completion as an optimization-based inpainting problem fortified by 2D StyleGAN and 3D geometry priors.

\noindent{\bf Unsupervised 3D Face Model :} There have been some studies to build 3d face model directly from 2D images such as \cite{tran2019learning} which learns a non-linear model from in-the-wild images and \cite{tewari2019fml} learns a complete model from videos. As a side-product of this approach, we attempt to build a texture model from a set of complete texture UV-maps of 2D images and compare it to GANFit~\cite{gecer2019ganfit} model which is trained by $\sim$ 10,000 high-quality 3D textures.

\section{Unsupervised UV Completion}
The key insight of our work is to utilize 2D face generator networks and 3D geometry in a progressive one-shot optimization procedure for texture completion and frontalization. Basically, our approach rotates an input image in 3D and fill-in the unseen regions by reconstructing the rotated image in a 2D face generator, based on the visible parts. This 2D reconstruction is performed by an optimization in the latent space of the generator. Finally, textures acquired from these generations are collected progressively to build a coherent texture UV-map. In this section, we explain the details of our method.

\subsection{3DMM Fitting \& Input Texture Acquisition}
\label{sec:input_texture}

\begin{figure}
\begin{subfigure}{0.24\linewidth}
\captionsetup{justification=centering}
    \includegraphics[width=\linewidth]{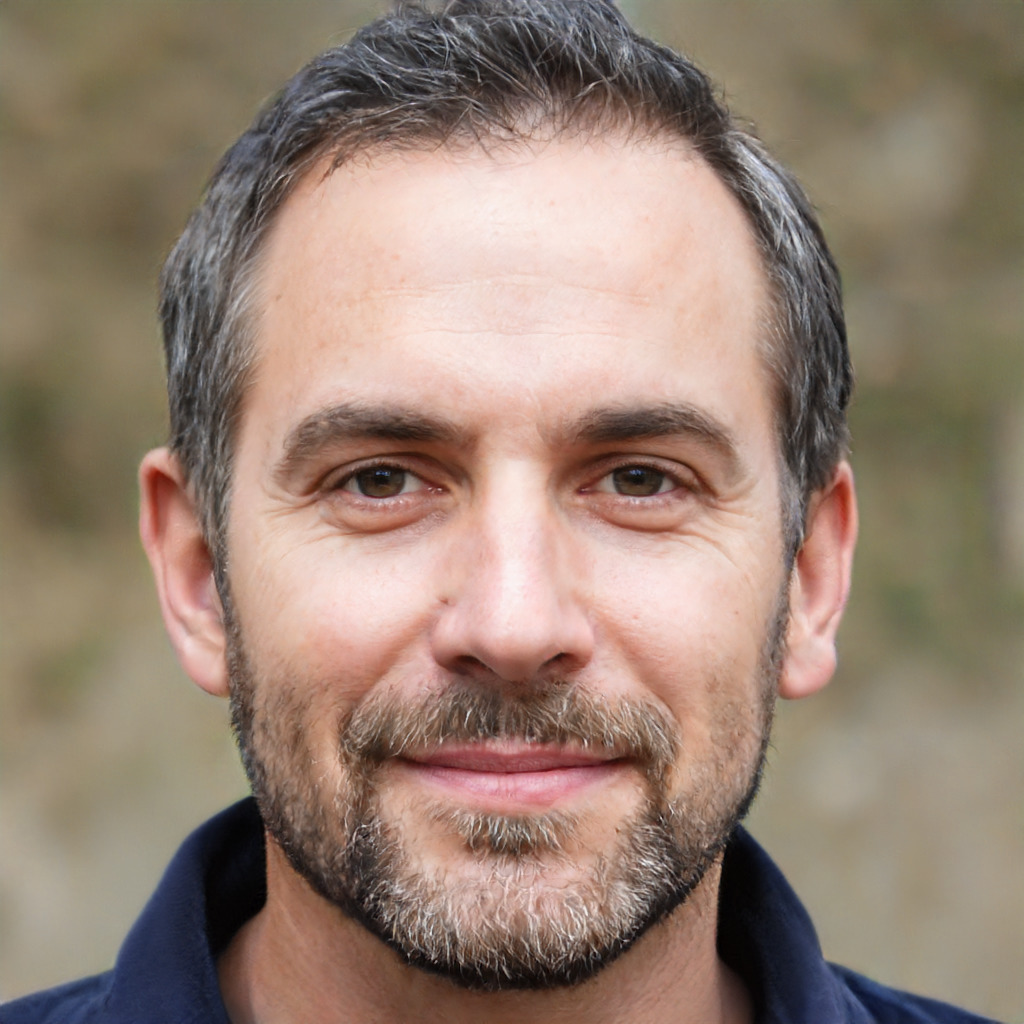}
    \vspace{-0.5cm}
    \caption{Input image ($\mathbf{I}_0$)}
    \label{fig:3DMM_input}
\end{subfigure}
\begin{subfigure}{0.24\linewidth}
\captionsetup{justification=centering}
    \includegraphics[width=\linewidth]{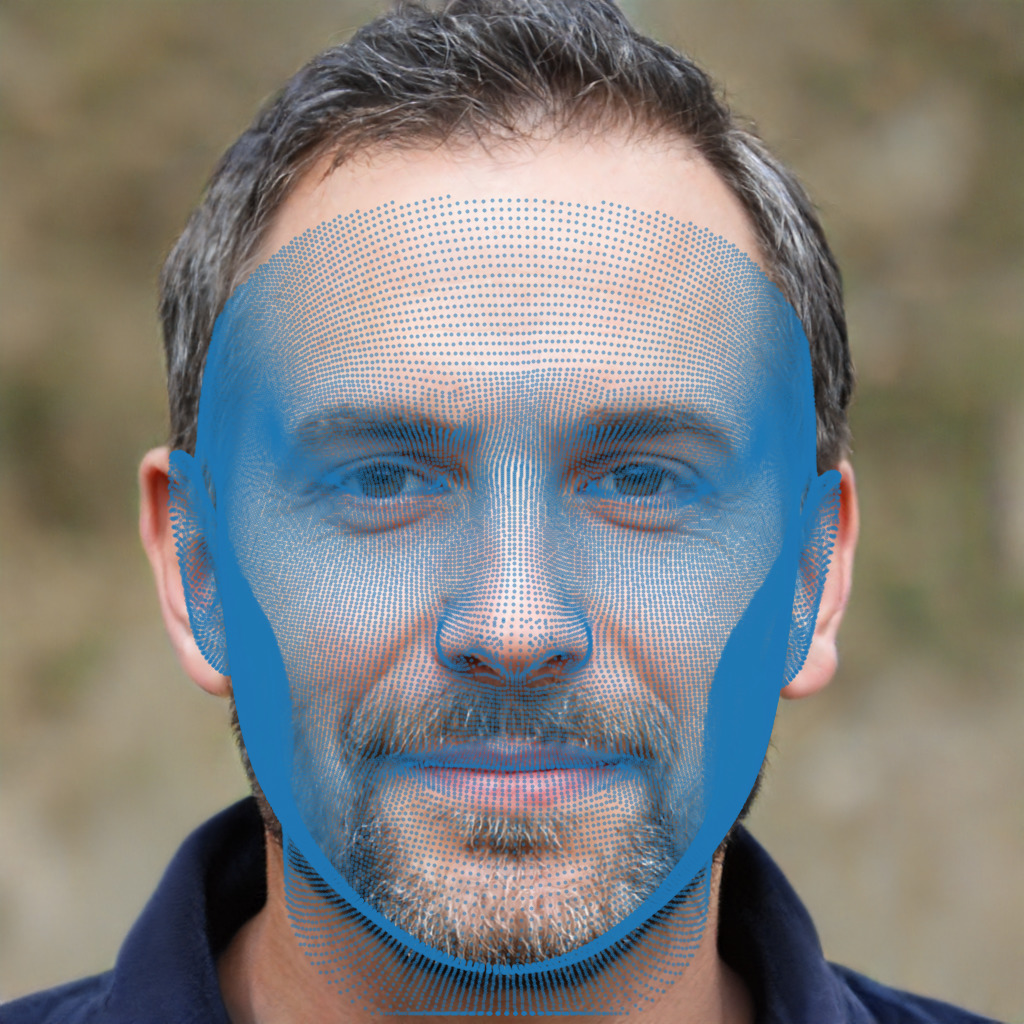}
    \vspace{-0.5cm}
    \caption{3D recons-\\truction($\mathbf{S'}$)}
    \label{fig:3DMM_landmarks}
\end{subfigure}
\begin{subfigure}{0.24\linewidth}
\captionsetup{justification=centering}
    \includegraphics[width=\linewidth]{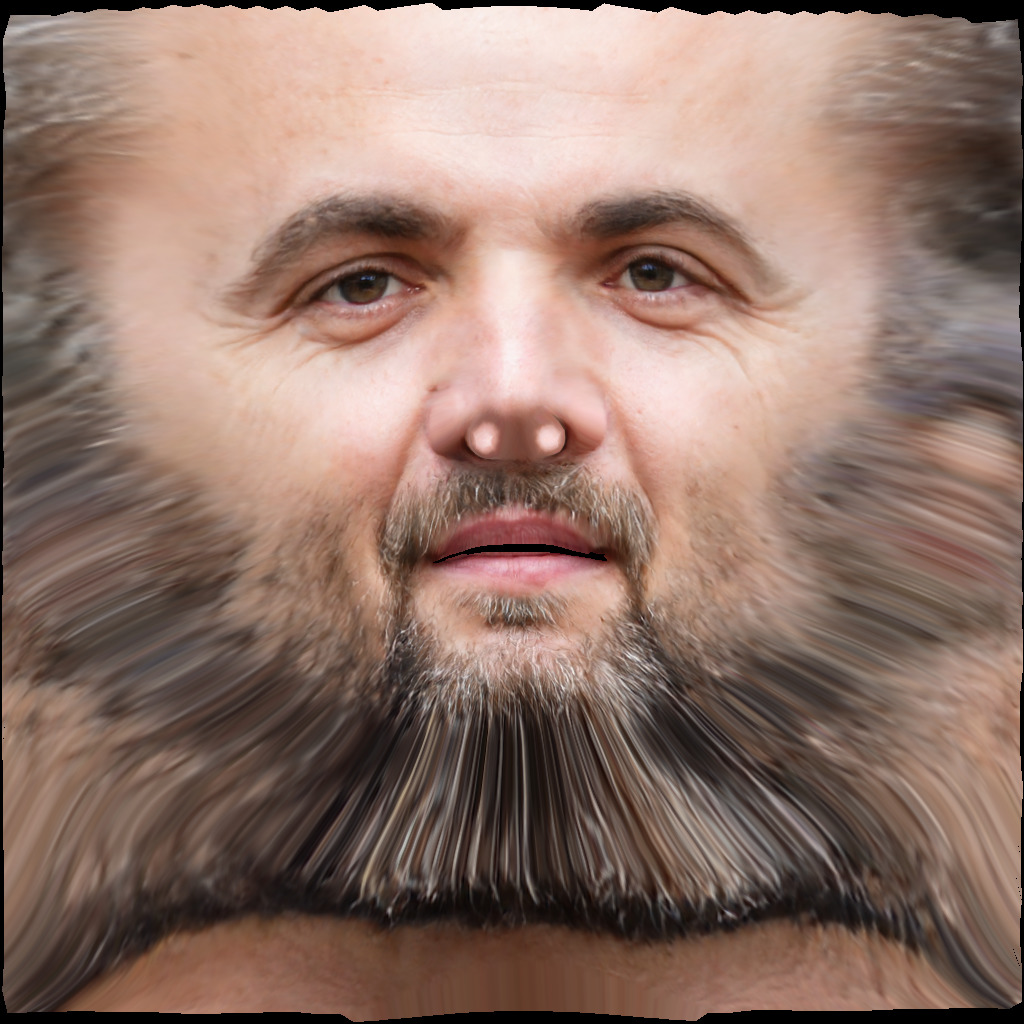}
    \vspace{-0.5cm}
    \caption{Interpolated UV map($\mathbf{T}_0$)}
    \label{fig:3DMM_uv}
\end{subfigure}
\begin{subfigure}{0.24\linewidth}
\captionsetup{justification=centering}
    \includegraphics[width=\linewidth]{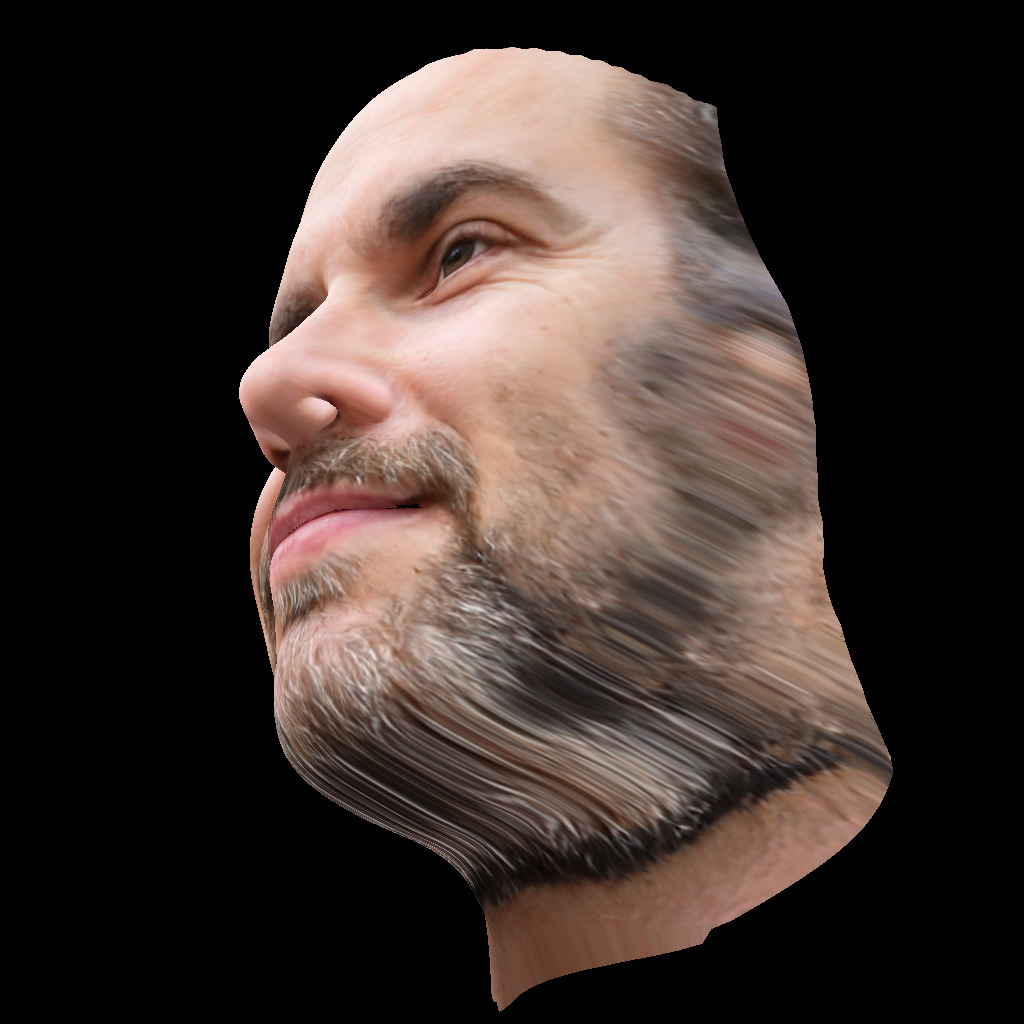}
    \vspace{-0.5cm}
    \caption{A different view ($\mathbf{I}_i$)}
    \label{fig:3DMM_rendered}
\end{subfigure}
\caption{3DMM Fitting and texture completion by nearest-neighbour interpolation. As can be seen in (d), interpolation method produces artefacts for different camera views.}
\label{fig:3DMM}
\end{figure}

For a given 2D face image $\mathbf{I}_0$, our approach relies on a rough estimation of its dense landmarks by a 3D reconstruction method. Therefore, we begin by fitting an \textit{off-the-shelf} 3DMM algorithm to estimate its geometry\footnote{Please note that no texture reconstruction from the 3DMM fitting algorithm is passed to the next stages.}  $\mathbf{S} \in \mathbb{R}^{n \times 3}$ and camera parameters $\mathbf{c}=[f,r_x,r_y,r_z,t_x,t_y,t_z]$. Let us define a 2D projection operation by a pinhole camera model with the function $\mathcal{P}(\mathbf{S}, \mathbf{c}) : \mathbb{R}^{n\times3}, \mathbb{R}^{7} \rightarrow \mathbb{R}^{n\times2}$, the geometry is then projected onto 2D image plane, \ie dense landmarks, by $
\mathbf{S'} = \mathcal{P}(\mathbf{S}, \mathbf{c})$.

Traditionally, high-quality 3D texture information can be stored in UV maps which assign 3D texture data into 2D planes with a universal per-pixel alignment for all textures. Each vertex of the geometry has a texture coordinate $t_{coord} \in \mathbb{R}^{n\times2}$ in the UV image plane in which the texture information is stored. In our approach, starting from the texture available in the input image, we progressively complete the texture in the UV space.

Given a set of 2D vertex coordinates, a texture UV map $\mathbf{T} \in \mathbb{R}^{w \times h \times 3}$ and texture coordinates, one can render a textured geometry by performing rasterization with barycentric interpolation expressed as $\mathcal{R} : (\mathbb{R}^{n\times2}, \mathbb{R}^{w \times h \times 3}, \mathbb{R}^{n\times2}) \rightarrow \mathbb{R}^{w' \times h' \times 3}$.

In order to acquire the visible part of the texture from the input image ($\mathbf{I}_0$), we perform a similar rendering by swapping vertex coordinates with texture coordinates and the texture UV map with the input image (\ie, image-to-UV rendering). In other words, the dense landmarks ($\mathbf{S'}$) from 3DMM fitting replace texture coordinates where the texture is actually the original image ($\mathbf{I}_0$). So, we unfold the input image into the UV space by giving the actual $t_{coord}$ of our topology as the vertex coordinates to be rendered. Consequently, the rendering is performed by the following:
\begin{align}
    \mathbf{T}_0 = \mathcal{R}'(t_{coord}, \mathbf{I}_0, \mathbf{S'})
\end{align}
in which image-to-UV rendering ($\mathcal{R}'$) is essentially same operation as UV-to-image rendering ($\mathcal{R}$), however, we denote them differently to avoid confusion.

An obvious motivation of this work can be seen in the illustration of this operation in Fig.~\ref{fig:3DMM}. After acquisition of the visible texture from the input image, we can see huge artefacts at invisible and narrow-angled parts of the geometry. Therefore, we explain how to detect and inpaint these regions by slowly building on top of the visible texture from the input image.

\subsection{Re-Rendering of the Mesh}
In order to fill-in the less-visible parts of the texture acquired from the original image, we rotate and render the fitted mesh by certain angles. We take the textured geometry as described in Sec.~\ref{sec:input_texture} and render it with a set of predefined camera parameters. The perspectives of these novel views are defined to maintain best visibility of every part of the face with a near-perpendicular view.

Given $\mathbf{c}_i$ ($i>0$) as the \textit{i}th novel camera parameters, we project geometry to the image plane and render texture geometry under this new perspective by the followings\footnote{The term $\overline{\mathbf{T}}_{i-1}$ refer to progressive texture of the previos iteration. It is explained in Sec.~\ref{sec:progressive}}:
\begin{align}
    \mathbf{S'}_i &= \mathcal{P}(\mathbf{S}, \mathbf{c}_i) \\
    \mathbf{I}_i &= \mathcal{R}(\mathbf{S'}_i, \overline{\mathbf{T}}_{i-1}, t_{coord})
    \label{eq:render}
\end{align}

\subsubsection{Building a Visibility Index}
Each of the novel perspectives dominates certain part of the texture map in terms of clarity and visibility, \ie bottom view is best for under-chin and side views are for cheeks. This visibility score can be defined in terms of the angle between the normal of each triangle and its vector pointing towards the camera. Meaning that, the acquired texture would have higher resolution and less artefact with lower angles between the two vectors, \ie for triangles that are facing towards the camera. For each perspective ($\mathbf{c}_i$), we extract a visibility UV map $\mathbf{V}_i$, ranging between $(-1,1)$ where $1$ indicates that the triangles around the vertex are facing towards the camera in average and $-1$ is facing the opposite direction. This process can be formulated by applying camera $\mathbf{c}_i$ to the geometry $\mathbf{S}$, and taking a dot product between vertex coordinates with respect to the camera and vertex normals.
%\mathbf{S}_i &= \mathcal{P}(\mathbf{S}, \mathbf{c}_i) \\
\begin{align}
\mathbf{V}_i &= \text{diag}( \frac{[\mathbf{S}'_i, \mathbf{h}]}{||[\mathbf{S}'_i, \mathbf{h}]||_2} \cdot \mathcal{N}(\mathbf{S}_i)^T)
\end{align}
where $\mathbf{h} \in \mathbb{R}^{n \times 1}$ stands for a vector of ones to make $\mathbf{S}'_i$ homogeneous. And $\mathcal{N}$ denotes the calculation the normals of the vertices. Some visibility score UV maps can be seen in Fig.~\ref{fig:angles} for different camera settings. Fig.~\ref{fig:angles_index} illustrates dominance map of all visibility scores, which we call \textit{visibility index} and use it for stitching texture maps that are generated from the optimization of different views. The binary masks of visibility index can be formulated as the following:
\begin{align}
\overline{\mathbf{V}}_i =   \bigcap\limits_{i \ne j} (\mathbf{V}_i > \mathbf{V}_j) \
\end{align}

\def \var {0.240}
\begin{figure}
    \begin{subfigure}{\var\linewidth}
        \includegraphics[width=\linewidth]{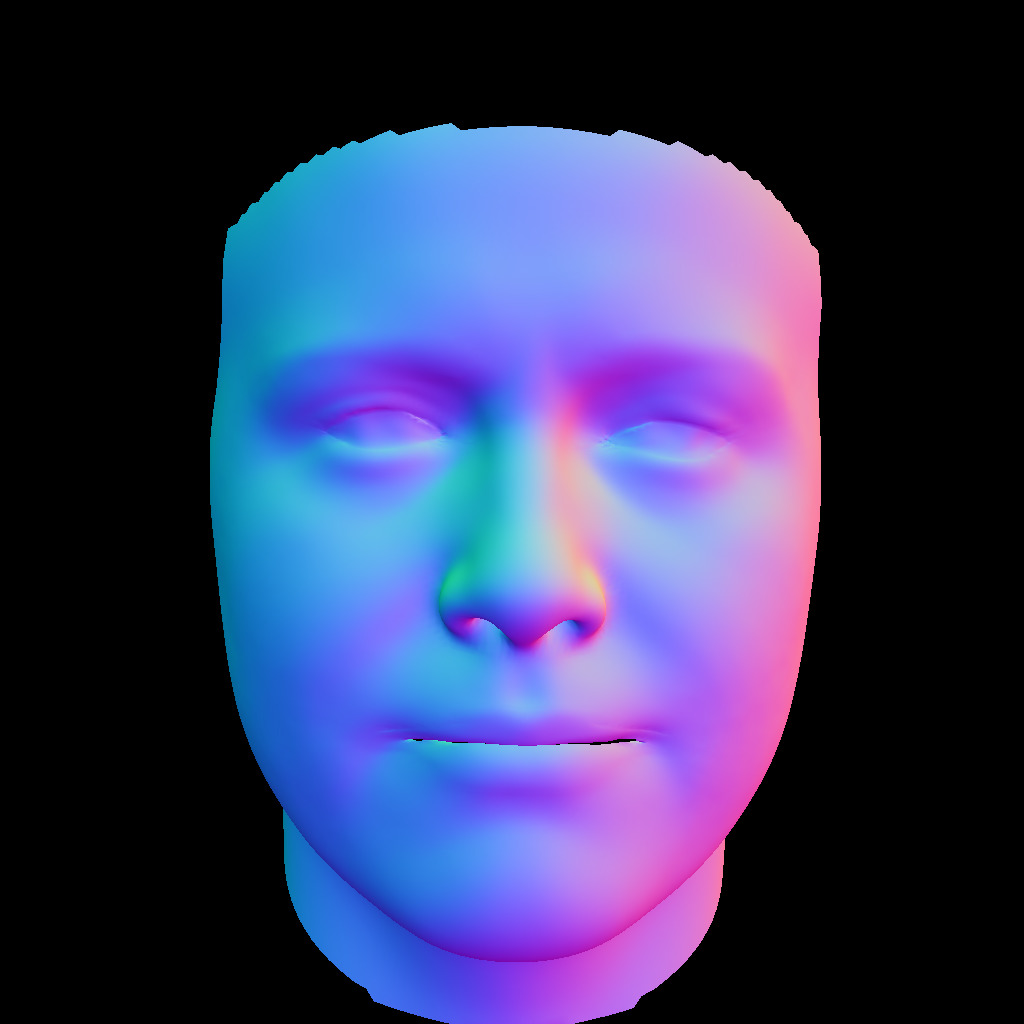}
        \label{fig:angles_normals}
        \vspace{-0.5cm}
        \caption{Input view}
    \end{subfigure}
    \begin{subfigure}{\var\linewidth}
        \includegraphics[width=\linewidth]{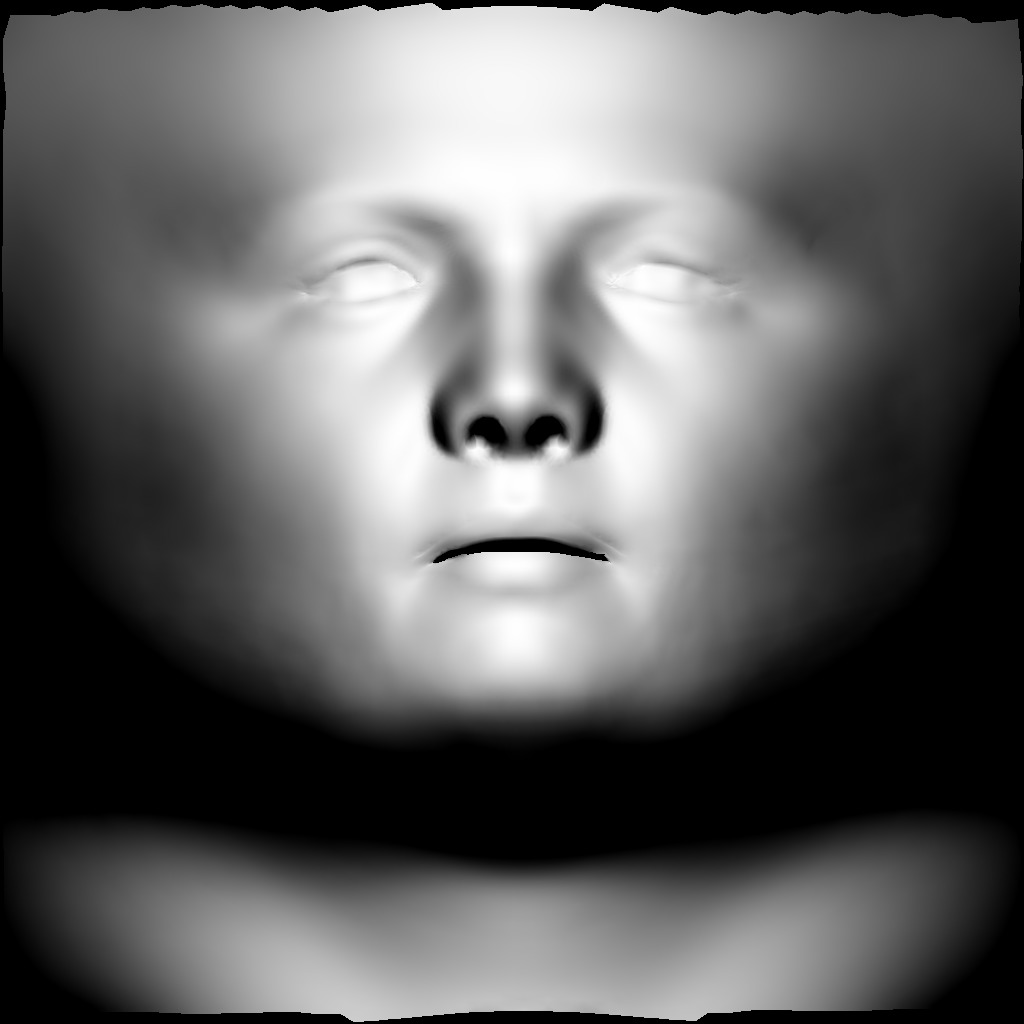}
        \label{fig:angles_input}
        \vspace{-0.5cm}
        \caption{Input view}
    \end{subfigure}
    \begin{subfigure}{\var\linewidth}
        \includegraphics[width=\linewidth]{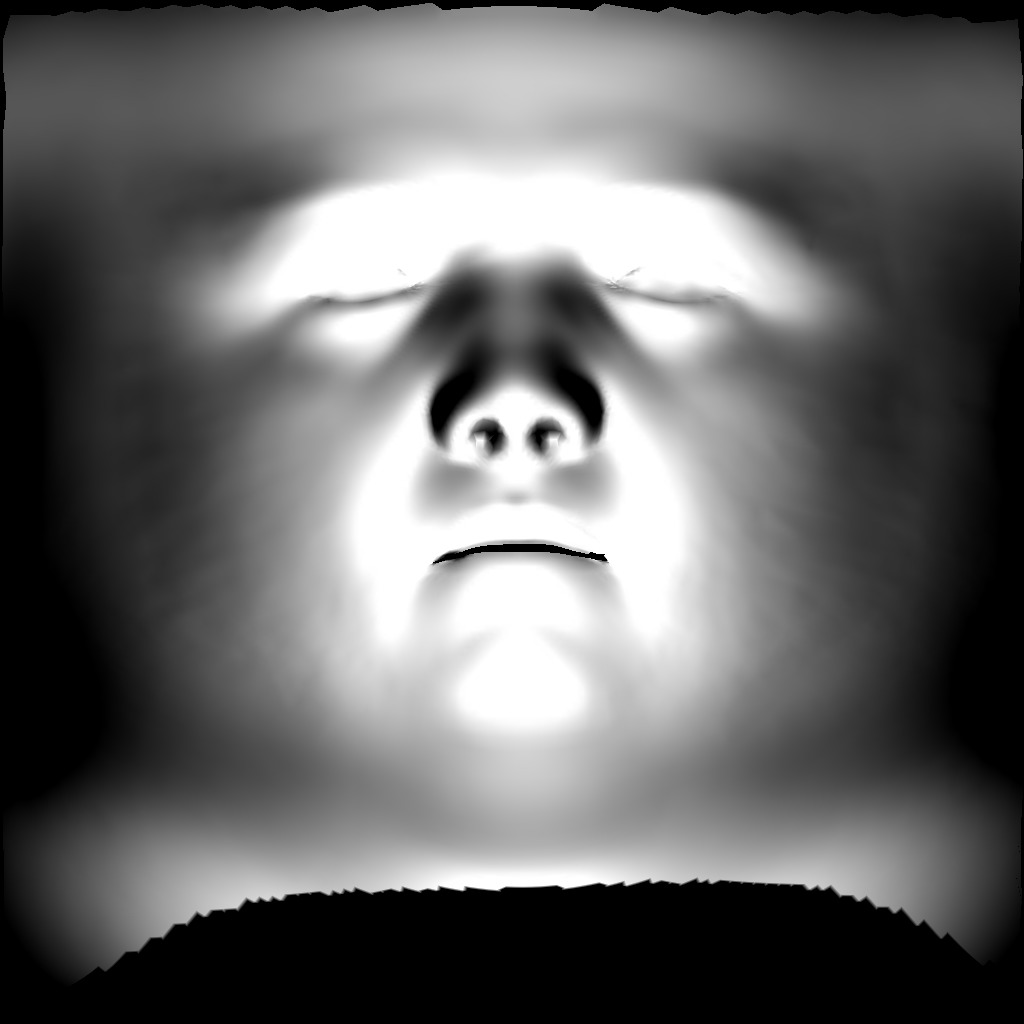}
        \label{fig:angles_bottom}
        \vspace{-0.5cm}
        \caption{Bottom}
    \end{subfigure}
    \begin{subfigure}{\var\linewidth}
        \includegraphics[width=\linewidth]{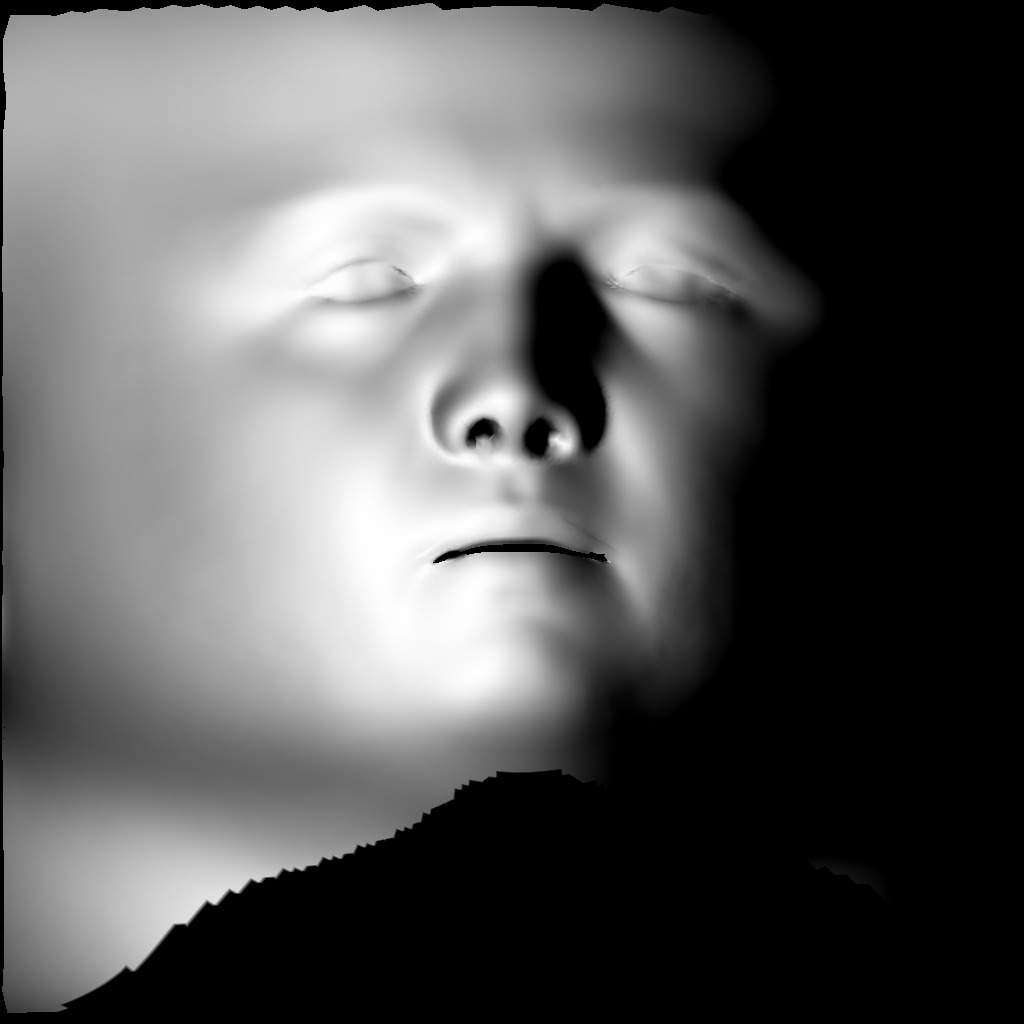}
        \label{fig:angles_bottom_left}
        \vspace{-0.5cm}
        \caption{Bottom-left}
    \end{subfigure}
    \begin{subfigure}{\var\linewidth}
    \includegraphics[width=\linewidth]{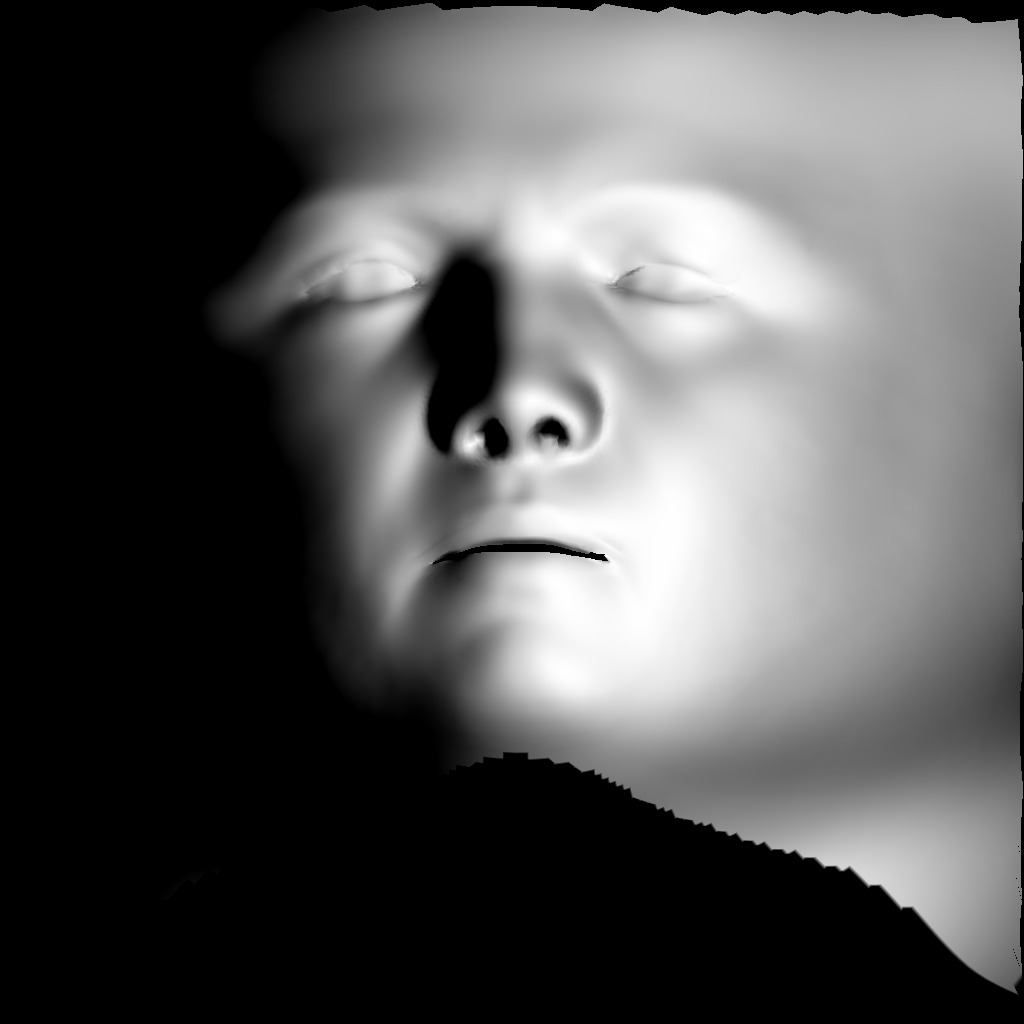}
        \label{fig:angles_bottom_right}
        \vspace{-0.5cm}
        \caption{$\!$Bottom-right$\!$}
    \end{subfigure}
    \begin{subfigure}{\var\linewidth}
        \includegraphics[width=\linewidth]{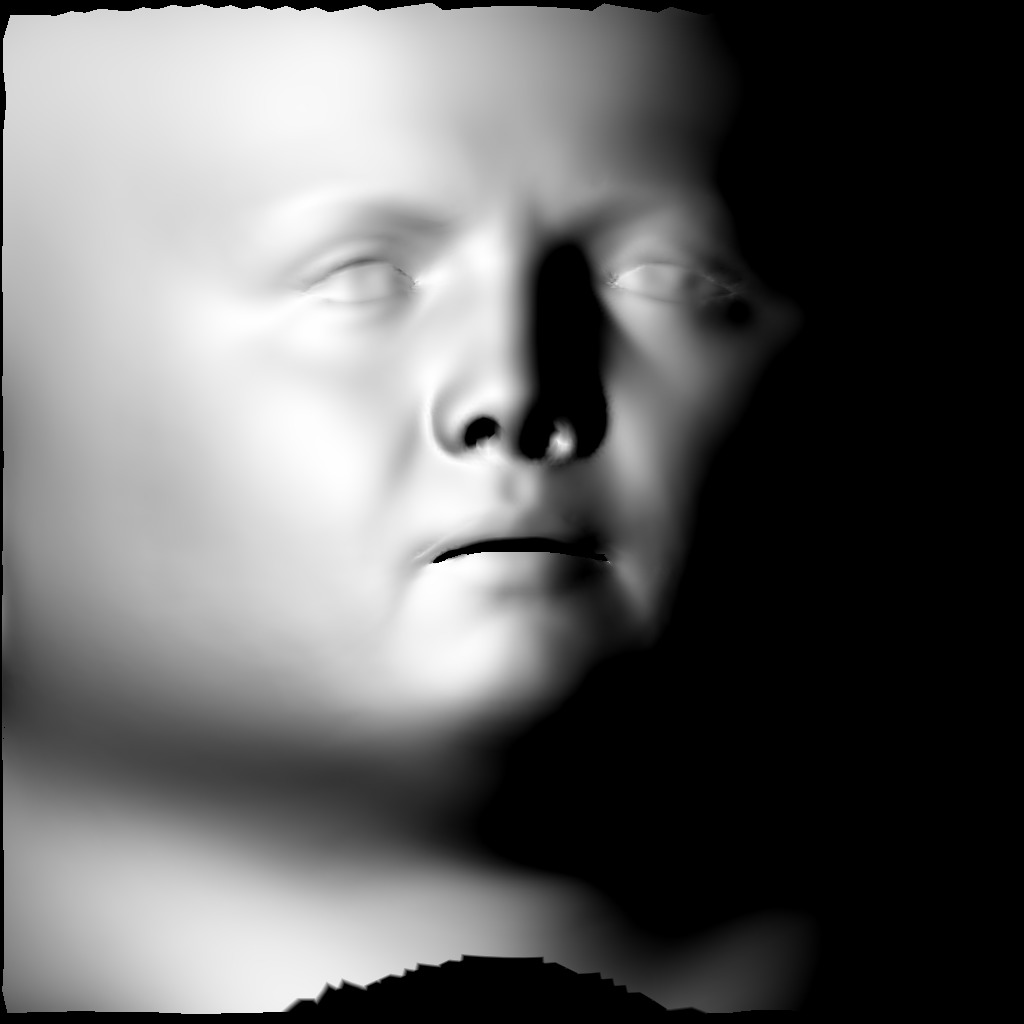}
        \label{fig:rangles_left}
        \vspace{-0.5cm}
        \caption{Left}
    \end{subfigure}
        \begin{subfigure}{\var\linewidth}
        \includegraphics[width=\linewidth]{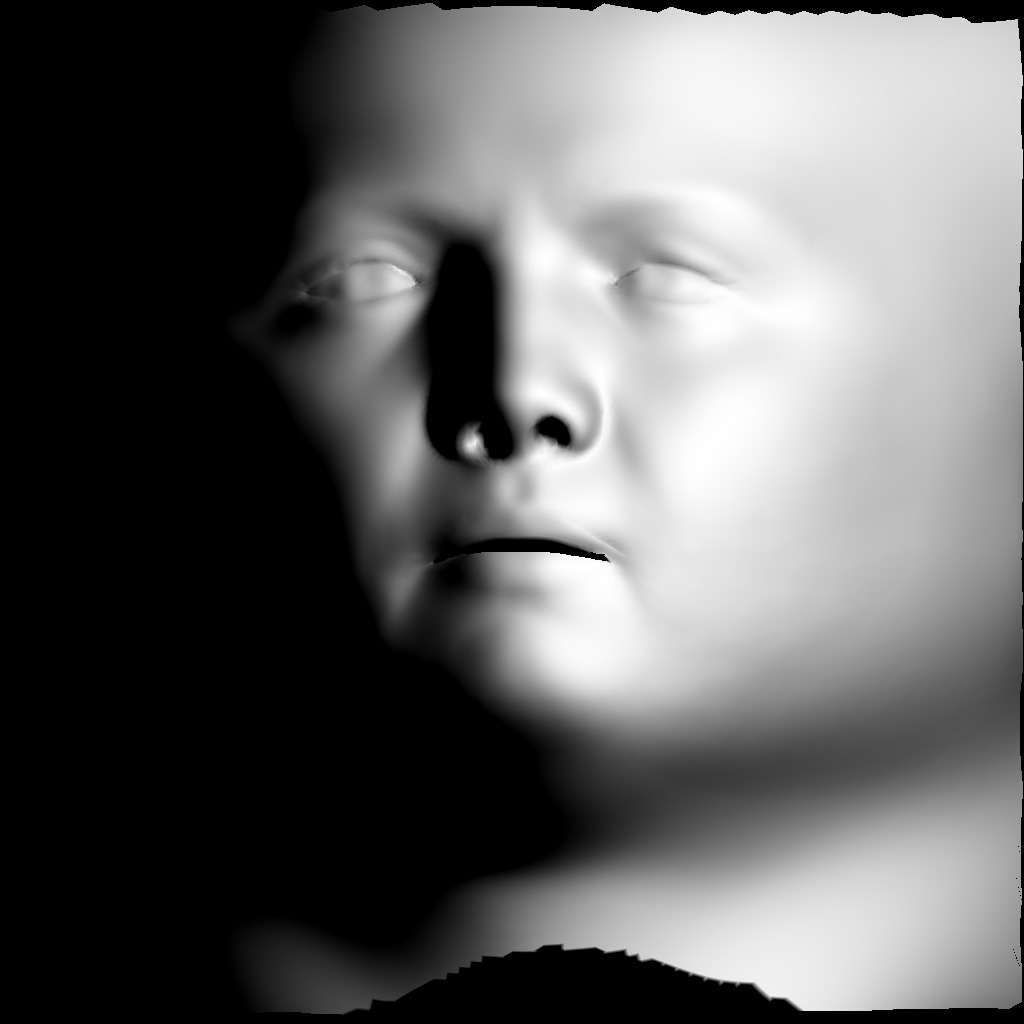}
        \label{fig:angles_right}
        \vspace{-0.5cm}
        \caption{Right}
    \end{subfigure}
        \begin{subfigure}{\var\linewidth}
        \includegraphics[width=\linewidth]{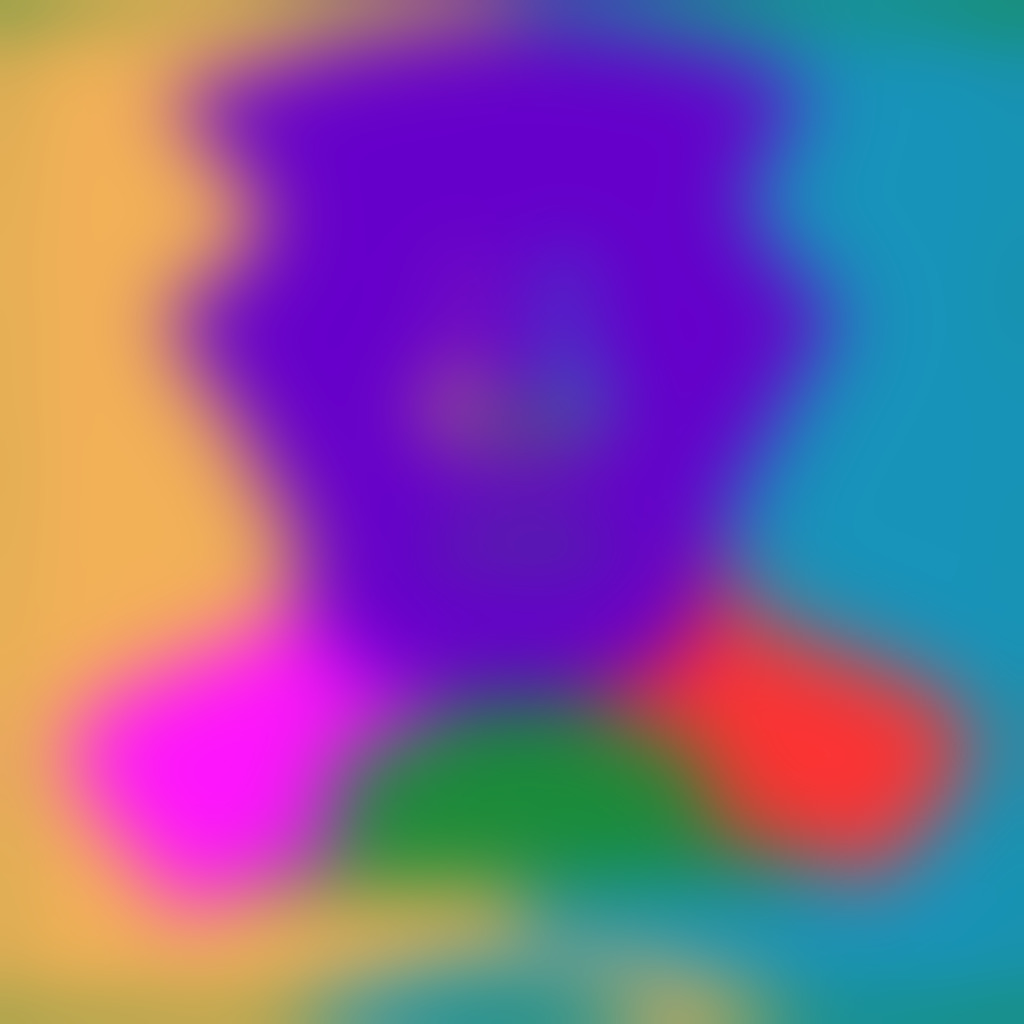}
        \vspace{-0.5cm}
        \caption{$\!$Visibility Ind.$\!\!\!\!$}
        \label{fig:angles_index}
    \end{subfigure}
    \caption{Visibility scores are to measure optimal camera angles with respect to facial surface in UV-map.(a) $\mathbf{V}_0$. of input image (b-g) $\mathbf{V}_i$ of different views. (h) Visibility index ($\overline{\mathbf{V}}_i$): an index of optimal angles for texture acquisition.}
    \label{fig:angles}
\end{figure}

\subsection{Inpainting by Projection}

\begin{figure*}[t]
\centering
    \includegraphics[width=1.0\linewidth]{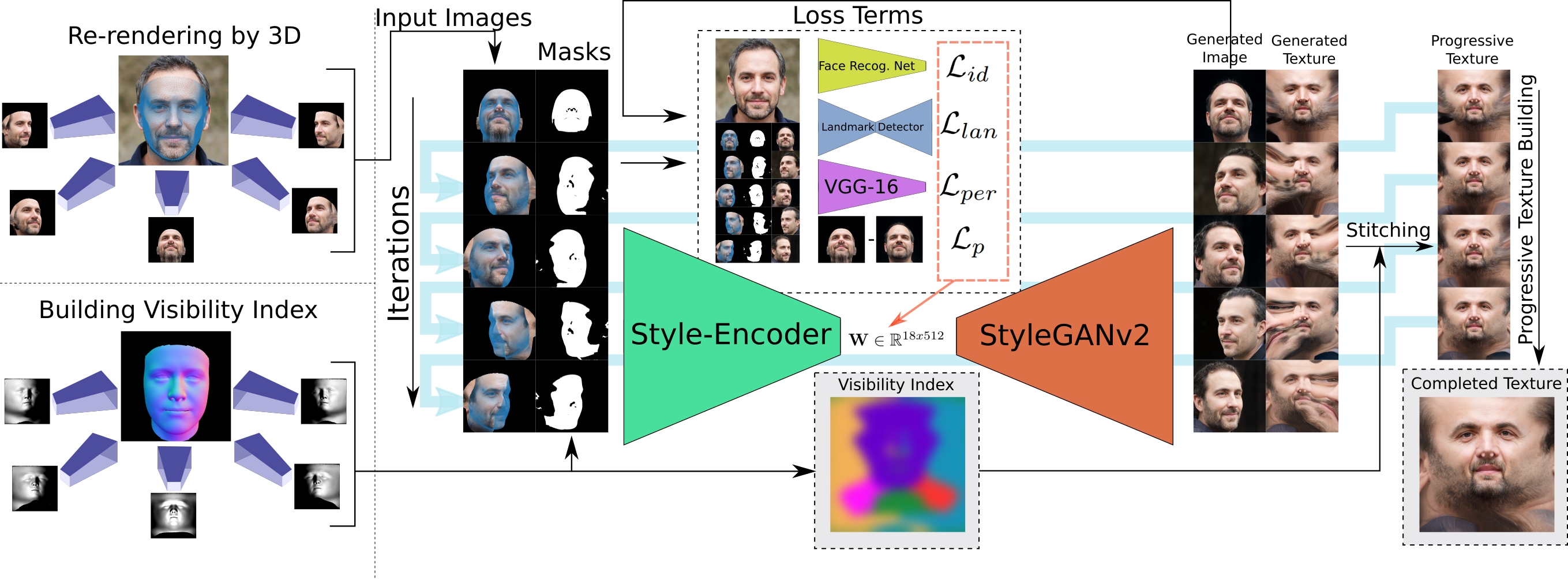}
    \caption{Overview of the method. The proposed approach iteratively optimizes the texture UV-maps for different re-rendered images with their masks. At the end of each optimization, generated images are used to acquire partial UV images by dense landmarks. Finally, the completed UV images are fed to the next iteration for progressive texture building.}
    \label{fig:overview}
\end{figure*}

The main assumption of this work is that we can utilize a generator network trained by 2D images as a prior appearance model for inpainting. Since we extracted a part of texture from the original image in Sec.~\ref{sec:input_texture}, we can now use it for conditional projection to styleGAN model to generate high quality and consistent faces for the invisible part.

\subsubsection{Masking}
In order to separate visible and invisible regions, for each novel view, binary masks are generated from the visibility scores ($\mathbf{V}_i$). We empirically found that intersection of two masks gives the best results: 1) regions where the visibility score of the original camera is higher than a threshold ($\mathbf{V}_0 > t_1$), and 2) regions where the visibility score of the original camera perspective is higher than the target camera\footnote{Other cameras are handicapped by a factor of 2 to enlarge texture from the original image}. Additionally, we progressively enlarge this mask by the dominant regions of all previously processed camera views, which is explained in Sec.~\ref{sec:progressive}.

The mask as explained above would give a UV mask which is then rendered by the current camera parameters $\mathbf{c}_i$ (\ie similar to Eq.~\ref{eq:render}). The whole mask extraction process can be formulated as the following:
\begin{align}
    \mathbf{M}^{UV}_i &=\big((\mathbf{V}_0 > t_1) \cap (2\mathbf{V}_0 > \mathbf{V}_i)\big) \cup \displaystyle\bigcup\limits_{i>j} \overline{\mathbf{V}}_j\\
    \mathbf{M}_i &= \mathcal{R}(\mathbf{S'}_i, \mathbf{M}^{UV}_i, t_{coord})
\end{align}

\subsubsection{Face Generation}
The proposed approach requires a good quality generator that can synthesize face images from an arbitrary noise vector. Therefore, we borrow one of the \textit{state-of-the-art} GAN network: StyleGANv2~\cite{karras2020analyzing} for this task. The StyleGAN or StyleGANv2 generators are particularly practical for this task as they consist of a mapping network that adds flexibility for manipulation and better projection. The mapping network ($\mathcal{G}_M: \mathbb{R}^{1 \times 512} \rightarrow \mathbb{R}^{18 \times 512}$) inputs a noise vector $\mathbf{z} \in \mathbb{R}^{1 \times 512}$ and generates an extended latent parameters $\mathbf{W} \in \mathbb{R}^{18 \times 512}$. The generator network can synthesize face images from this extended latent parameters fed into its different layers, \ie $\mathcal{G}: \mathbb{R}^{18 \times 512} \rightarrow \mathbb{R}^{h' \times w' \times 3}$. In this work, we optimize only based on $\mathbf{W}$ which we call latent parameters and ignore the mapping network.

During the forward pass of the optimization, we generate an image $\mathbf{G}_i^*$ by the generator network $\mathcal{G}(\mathbf{W}_i^*)$ and extract a set of features for the energy terms that we explain below. As explained in Sec.~\ref{sec:optimization}, the loss is backpropagated to find a good generation by updating the latent parameters $\mathbf{W}$.

\subsubsection{Energy Functions}

\paragraph{Photometric Loss :}
Obviously, one of the simplest form of supervision is photometric loss which  encourages low-level similarity at the visible part of the image. Although simpler form of photometric loss can be defined as pixel-wise mean absolute difference between two images, we empirically find that log-cosh loss provides smoother convergence. Log-cosh loss can be defined as the following:
\begin{align}
    \mathcal{L}_p = \frac{1}{w' \times h' \times 3} \sum^{w' \times h' \times 3} \log \Big( \cosh \big( \mathbf{M}_i \odot (\mathbf{I}_i - \mathbf{G}_i) \big)  \Big)
\end{align}
where $\odot$ stands for element-wise multiplication.
%However, we would like to optimize this loss term only based on the clearly visible part of the texture coming from the original image. Therefore, we filter this loss by the masks obtained in Sec.~, and consequently generator is relaxed for the invisible parts. Nevertheless,

\paragraph{Identity Loss :}
Since photometric loss is only concerned by the low-level similarity, it struggles to achieve smooth convergence. Following~\cite{gecer2019ganfit, cole2017synthesizing, genova2018unsupervised, gecer2018facegan}, we exploit identity features from a pretrained face recognition network~\cite{deng2018arcface} in order to capture good identity resemblance with the original image. Given a network $\mathcal{F}: \mathbb{R}^{h' \times w' \times c} \rightarrow \mathbb{R}^{512}$, we calculate the cosine distance between the identity features of the generated image and the input image as following:
\begin{align}
\mathcal{L}_{id} = 1 - \frac{\mathcal{F}(\mathbf{I}_0) \cdot \mathcal{F}( \mathbf{G}_i))}{||\mathcal{F}(\mathbf{I}_0)||_2 ||\mathcal{F}(\mathbf{G}_i))||_2}
\label{eq:id_loss}
\end{align}

\paragraph{Perceptual Loss :}
Following the previous studies~\cite{gecer2019ganfit,abdal2019image2stylegan}, we exploit high-level similarity features, known as perceptual loss, to regularize convergence. We empirically choose 9th layer of a VGG-16 network that is pretrained as an ImageNet classifier as below:
\begin{align}
    \mathcal{L}_{per} = \sum \log \bigg( \cosh \Big( \mathbf{M}_i \odot \big(\text{VGG}(\mathbf{I}_i) - \text{VGG}(\mathbf{G}_i) \big)\Big)\bigg)
\end{align}

\paragraph{Landmark Loss :}
All previous objectives are segmented by the visibility mask that covers the face partially. Therefore, invisible parts become totally relaxed, which leads to ill-aligned generations with the rendered dense landmarks ($\mathbf{S}'_i$). To this end, we propose to minimize the landmark distance between $\mathbf{I}_i$ and $\mathbf{G}_i$. As we rotate 3D mesh with a fixed topology, sparse landmark locations of the rendered images can be easily obtained from the mesh with pre-defined landmark indices ($l \in \mathbb{N}^{68}, l<n$), \ie  ($\mathbf{S}'_i(l)$).
In order to extract landmarks of the generated image ($\mathbf{G}_i$) during the optimization \footnote{In order to flow the gradient from landmark loss, landmarks need to be computed by differentiable connections. To the best of our knowledge, this is the first attempt of such point-based supervision to a 2D image.}, we employ a differentiable landmark estimator~\cite{deng2018cascade} defined as $\mathcal{K} : \mathbb{R}^{w' \times h' \times 3} \rightarrow \mathbb{R}^{68 \times 2}$. And the loss is expressed as:
\begin{align}
    \mathcal{L}_{lan} = \frac{1}{68}\sum^{68} ||\mathcal{K}(\mathbf{G}_i) - \mathbf{S}'_i(l) ||_2
\end{align}

\subsubsection{Projection}
\label{sec:optimization}

\paragraph{Initialization by Regression :}
Following~\cite{abdal2019image2stylegan}, we train a regressor CNN network $\mathcal{E}: \mathbb{R}^{h' \times w' \times 3} \rightarrow \mathbb{R}^{18 \times 512}$ from random styleGAN generated images ($\mathcal{G}(\mathcal{G}_M(\mathbf{z})), \mathbf{z}\sim\mathcal{N}(\mathbf{0},\mathbf{I})$) to predict their latent parameters ($\mathbf{W} = \mathcal{G}_M(\mathbf{z})$). We initialize $\mathbf{W}$ by the regression of this network for the rendered images, \ie $\mathbf{W}^* = \mathcal{E}(\mathbf{I}_i)$. Initializing the latent parameters with this regression not only accelerate the convergence but also assist optimizer to avoid local minimas.

\paragraph{Optimization :}
Given a rendered image $\mathbf{I}_i$, its respective mask $\mathbf{M}_i$, and dense landmarks $\mathbf{S}'_i$, our goal is to find the best latent parameters ($\mathbf{W}_i$) to reconstruct $\mathbf{I}_i$ by a pretrained StyleGANv2 generator $\mathcal{G}$. To this end, we first align $\mathbf{I}_i$, $\mathbf{M}_i$, and $\mathbf{S}'_i$ to the alignment template of StyleGANv2. And, we perform gradient descent optimization by ADAM optimizer~\cite{kingma2014adam} with a weighted sum of loss functions defined above:
\begin{equation}
\begin{aligned}
    \min_{\mathbf{W}_i} \mathcal{L}_{total} (\mathbf{W}_i) = \lambda_p \mathcal{L}_p + \lambda_{id} \mathcal{L}_{id} + \lambda_{per} \mathcal{L}_{per} +  \lambda_{lan} \mathcal{L}_{lan}
\end{aligned}
\label{eq:optimization}
\end{equation}

After convergence, we synthesize a face image with the novel view $\mathbf{c}_i$ by $\mathbf{G}_i = \mathcal{G}(\mathbf{W}_i)$. Finally, we can acquire partial texture in the same way as input texture acquisition in Sec.~\ref{sec:input_texture} by $\mathbf{T}_i = \mathcal{R}'(t_{coord}, \mathbf{G}_i, \mathbf{S'}_i)$.

\subsection{Progressive Texture Building for Consistency}
\label{sec:progressive}

In order to generate globally consistent texture maps, we run the optimization for each of the camera views \textit{iteratively} to progressively improve the texture UV map. After every iteration, we blend the generated UV map ($\mathbf{T}_i$) into the current UV map at the dominated pixels ($\overline{\mathbf{V}}_i$) by that particular camera settings $\mathbf{c}_i$.
\begin{align}
    \overline{\mathbf{T}}_i = \overline{\mathbf{V}}_i \odot \mathbf{T}_i +  (1 - \overline{\mathbf{V}}_i) \odot \overline{\mathbf{T}}_{i-1} 
\end{align}

\paragraph{Blending:} Texture UV-maps are stitched by alpha blending for smooth shift between different UV maps. Also, they are RGB normalized by Gaussian statistics at the intersection of visibility indices $\overline{\mathbf{V}}_0$ and $\overline{\mathbf{V}}_i$~\footnote{Normally, these two indices do not overlap, however we build $\overline{\mathbf{V}}_i$ without the handicap to find out true dominated regions. And $\overline{\mathbf{V}}_0$ is from the previous index in which it is given advantage by a factor of 2}.

\paragraph{Face Frontalization:} Finally, with the complete UV map $\overline{\mathbf{T}}$, we render it once more by a frontal camera and perform a final optimization as in Eq.~\ref{eq:optimization}
\label{sec:progressive} to generate the frontal image of the input image.

\section{Experiments}
We implement the proposed approach in Tensorflow framework~\cite{tensorflow2015-whitepaper} and it takes around 5 minutes to UV-complete and frontalize an input image. Unfortunately, some of the preprocessing steps are CPU-intensive, therefore is a room for further efficiency. We have used geometry fitting pipeline of GANFit~\cite{gecer2019ganfit} as a preprocessing step. Other than pretrained networks for the loss function, the method itself does not require any additional training data. In the following, we illustrate some qualitative and quantitative results of our method.

\begin{figure}[t]
\centering
\includegraphics[width=1.0\linewidth]{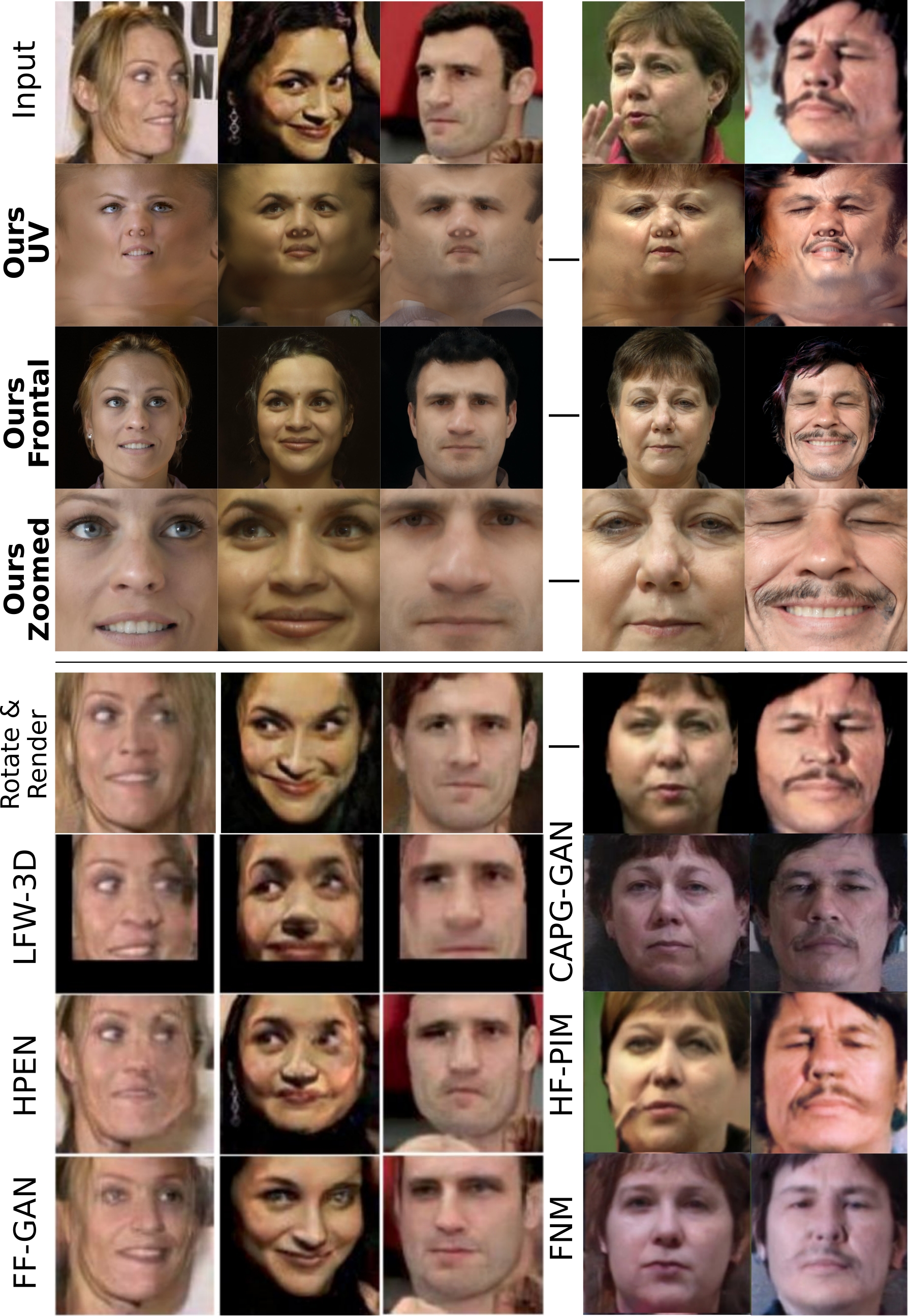}
\caption{Comparison of our frontalization to others: Rotate\&Render~\cite{zhou2020rotate}, FNM~\cite{qian2019unsupervised}, CAPG-GAN~\cite{hu2018pose}, FF-GAN~\cite{yin2017towards}, HF-PIM~\cite{cao2018learning}, HPEN~\cite{zhu2015high}, LFW-3D~\cite{hassner2015effective}}
\label{fig:comp2}
\end{figure}

\subsection{Unsupervised Texture Model: UTEM}

Many 3D texture reconstruction approaches rely on large-scale high-quality 3D appearance data which is costly to collect, difficult to maintain diversity (\eg ethnicity, age) and often kept private due licensing issues. On the other hand, large-scale high-quality 2D face datasets are widely available~\cite{lee2019maskgan,karras2017progressive} for all. As a by-product of our approach, we build a 3D texture model by completing texture UV-maps for $\sim$1,500 images from CelebA-HQ~\cite{lee2019maskgan} (as can be seen in Fig.~\ref{fig:utem} a), without any 3D data collection. After the completion, we train a GAN~\cite{karras2017progressive} as a GAN-based texture model and perform 3DMM fitting similar to \cite{gecer2019ganfit}. We call this model UTEM and show some generated samples in Fig.~\ref{fig:utem} b. The 3DMM fitting results by the original GANFit~\cite{gecer2019ganfit} and the one with UTEM texture model can be seen in the last two rows of Fig.~\ref{fig:comp1}. The reconstructed textures show similar identity recovery and quality as GANFit textures, and it will be available for all.
%~\footnote{we will make the code and the model public after publication.}

\subsection{Qualitative Results}
\begin{figure*}[t]
\centering
\vspace{-0.5cm}
\includegraphics[width=0.91\textwidth]{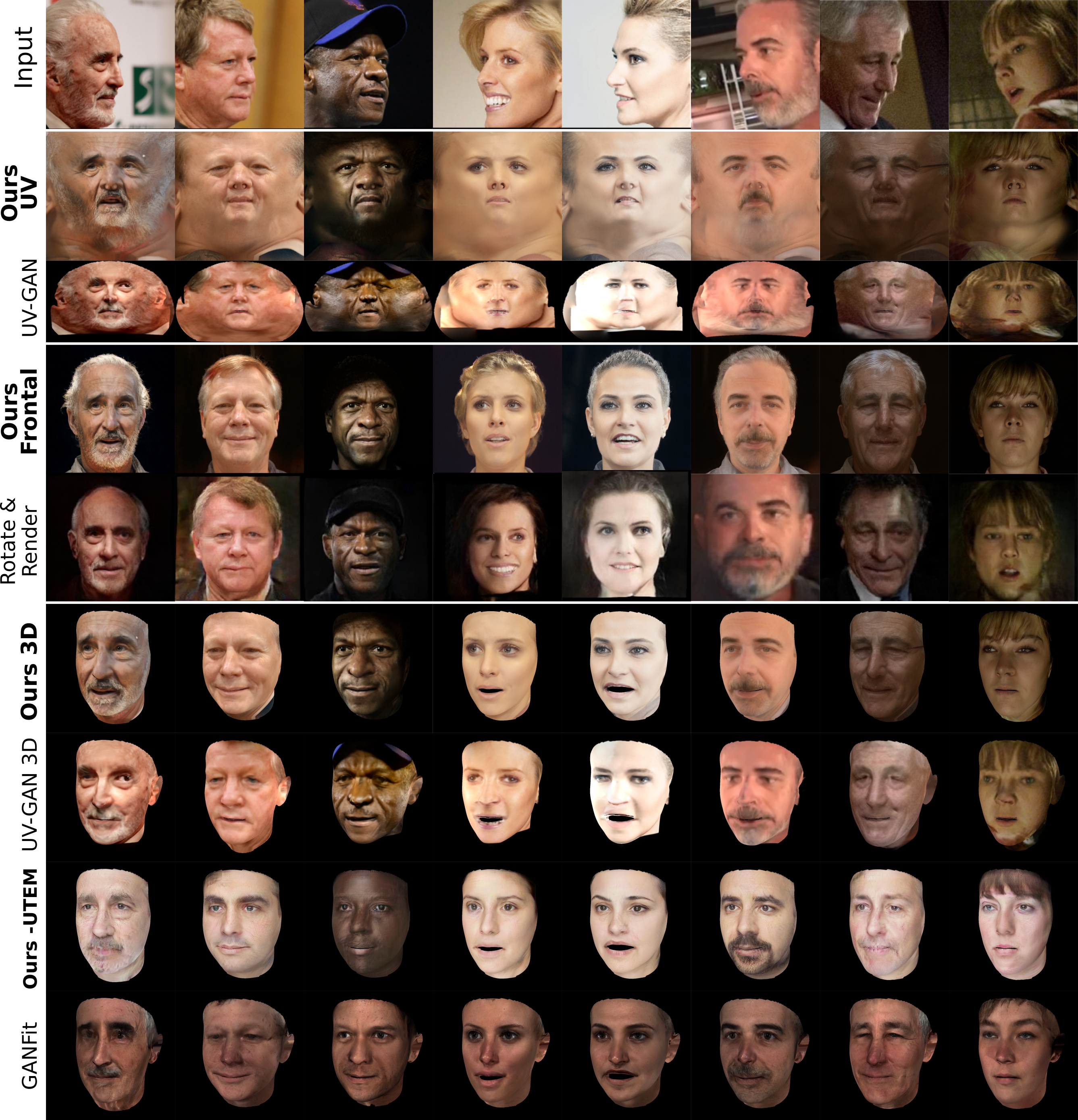}
\caption{Qualitative results in comparison with other state-of-the-art methods (UV-GAN~\cite{deng2018uv}, Rotate\&Render~\cite{zhou2020rotate} and GANFit~\cite{gecer2019ganfit}). (From Top to down) First block shows input images, second block UV-completion, third block frontalization, and the fourth block texture completion/reconstruction results.}
\label{fig:comp1}
\end{figure*}

We run our algorithm on some images in comparison with the recent state-of-the-art approaches, as shown in Fig.~\ref{fig:comp1},\ref{fig:comp2} and \ref{fig:teaser}. Fig.~\ref{fig:comp1} shows better quality and semantically meaningful UV-maps compared to UV-GAN~\cite{deng2018uv} and GANFit~\cite{gecer2019ganfit}. Frontalization results in both Fig.~\ref{fig:comp1} and \ref{fig:comp2} look superior to other previous methods in terms of identity-resemblance, artefacts and resolution.

%Benjamin_Netanyahu_0005
\def \var {0.17}
\begin{figure}
\foreach \j in {Julianne_Moore_0012}{
\foreach \i in {input,none,pixel,land,perc,all}{\includegraphics[width=\var\linewidth]{ablation/\i/\j_uv}}\\
\includegraphics[width=\var\linewidth]{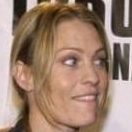}\foreach \i in {none,pixel,land,perc,all}{\includegraphics[width=\var\linewidth]{ablation/\i/\j_frontal}}\\}
\def \j {Robin_Wright_Penn_0001}
\foreach \i in {input,none,pixel,land,perc,all}{\includegraphics[width=\var\linewidth]{ablation/\i/\j_uv}}\\\begin{subfigure}{\var\linewidth}
\includegraphics[width=\linewidth]{ablation/input/\j_uv}
\tiny\caption{Input}\end{subfigure}\begin{subfigure}{\var\linewidth}
\includegraphics[width=\linewidth]{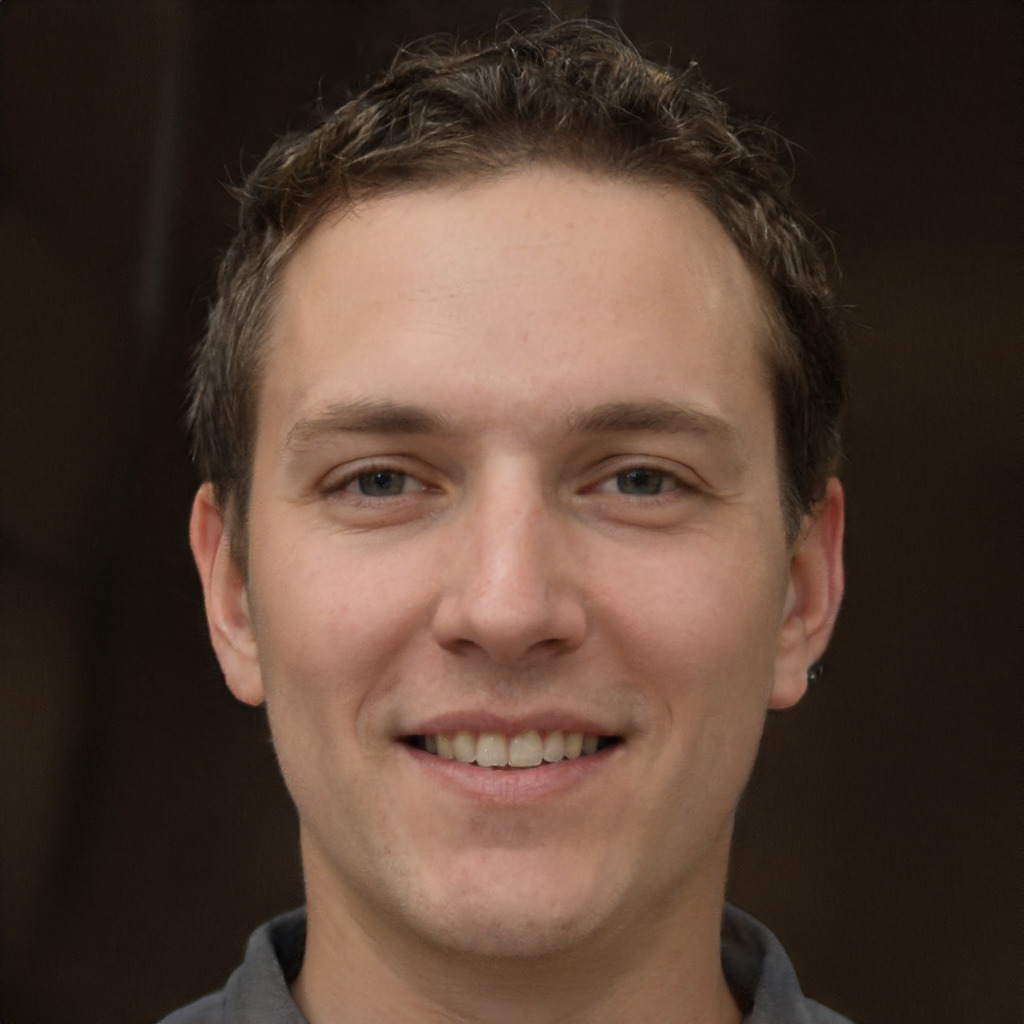}
\tiny\caption{$\mathcal{E}(\mathbf{I}_0)$}\end{subfigure}\begin{subfigure}{\var\linewidth}
\includegraphics[width=\linewidth]{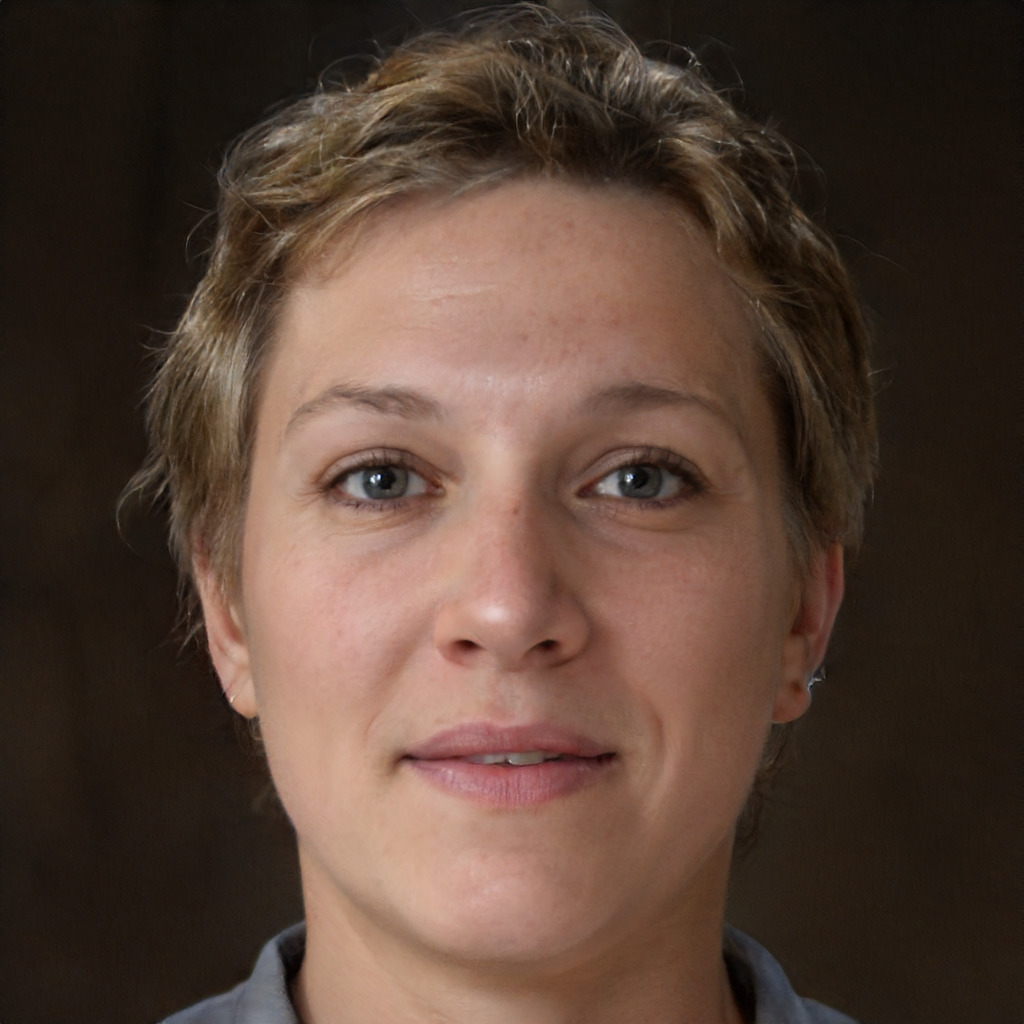}
\tiny\caption{+$\mathcal{L}_p$}\end{subfigure}\begin{subfigure}{\var\linewidth}
\includegraphics[width=\linewidth]{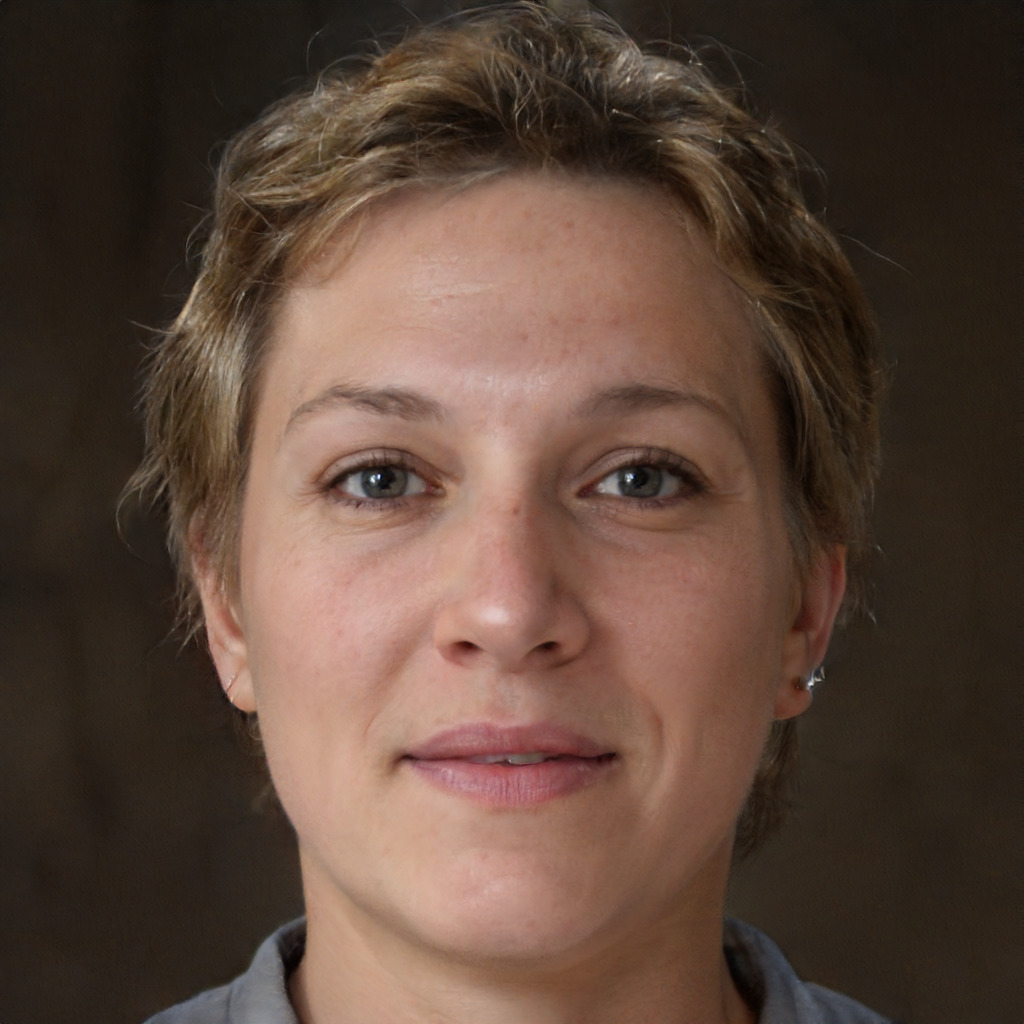}
\tiny\caption{+$\mathcal{L}_{lan}$}\end{subfigure}\begin{subfigure}{\var\linewidth}
\includegraphics[width=\linewidth]{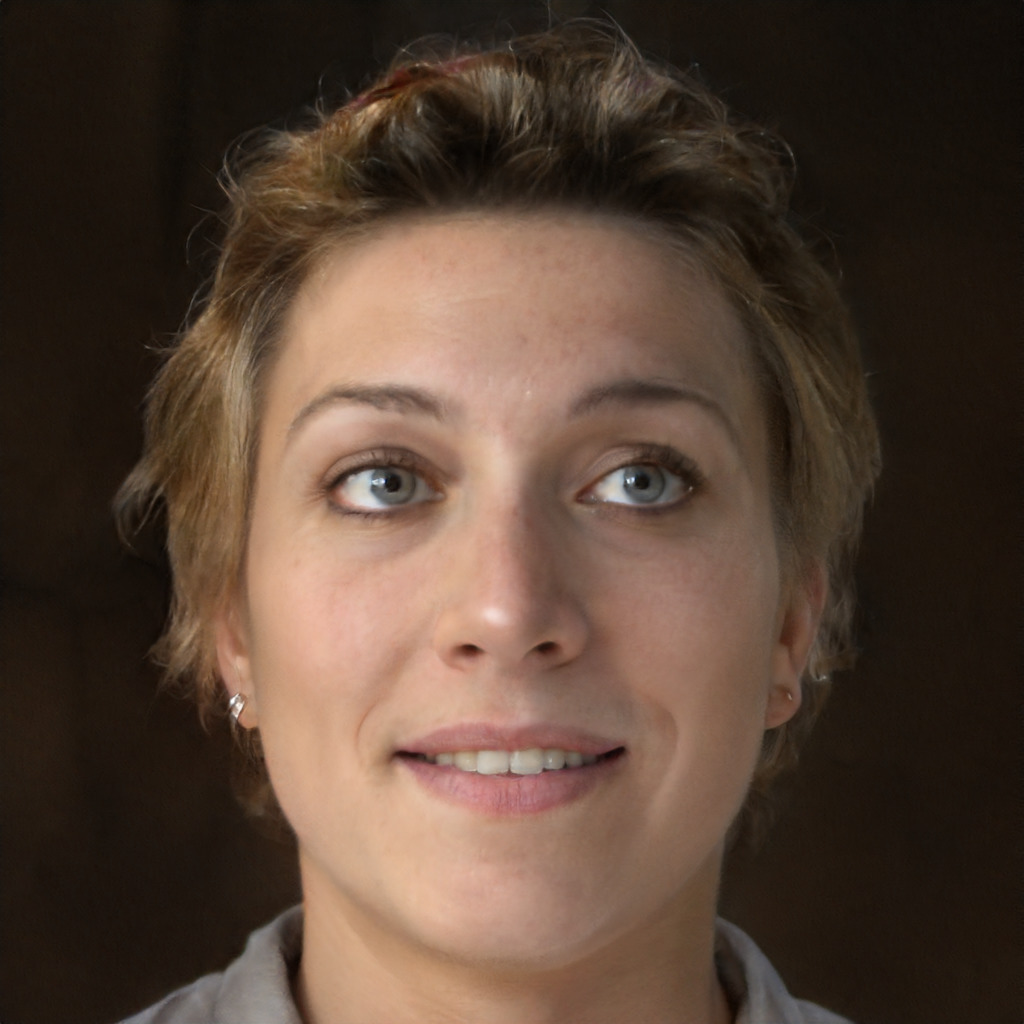}
\tiny\caption{+$\mathcal{L}_{per}$}\end{subfigure}\begin{subfigure}{\var\linewidth}
\includegraphics[width=\linewidth]{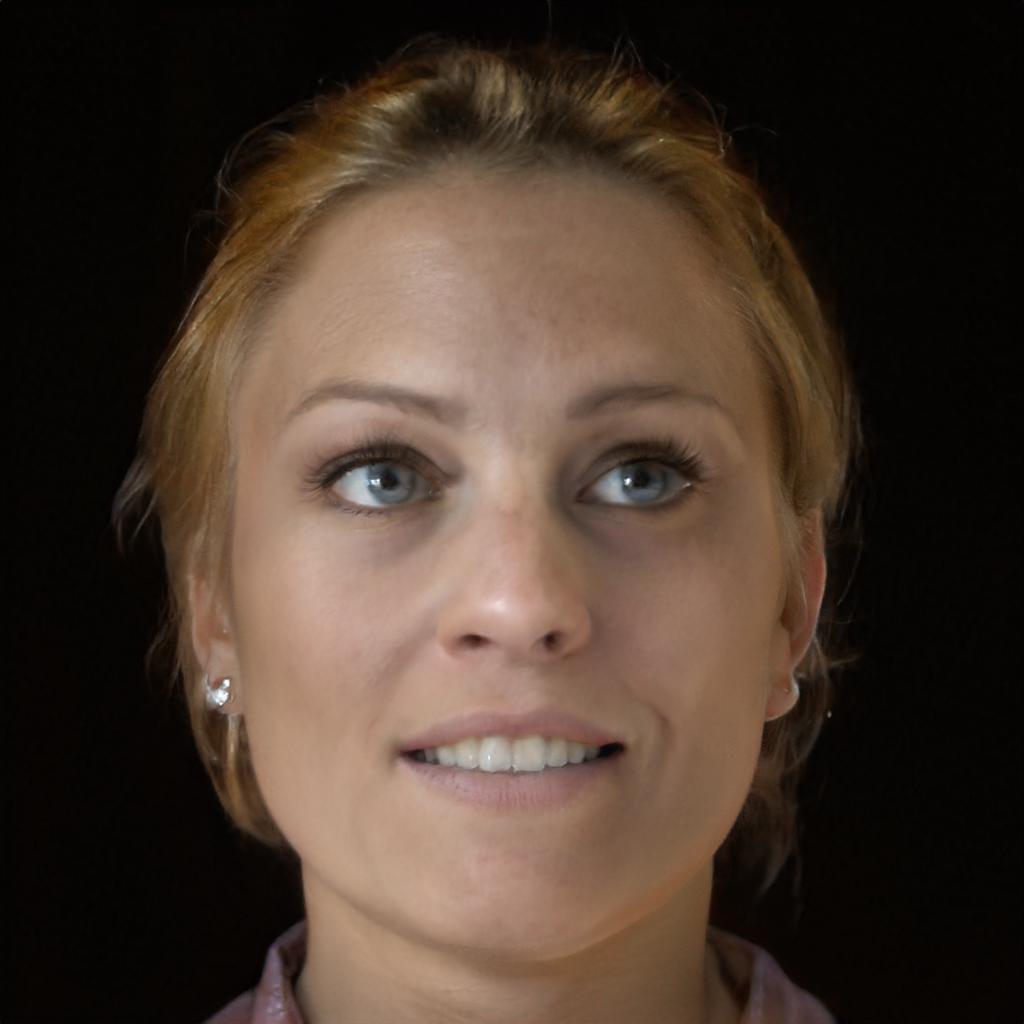}
\tiny\caption{+$\mathcal{L}_{id}$}\end{subfigure}
\caption{Additive ablation study. `+' refers to addition of that loss term compared to the column on the left. (f) refers to all loss term used in this paper, \ie \textbf{Ours}.}
\label{fig:ablation}
\end{figure}

\def \var {0.13}
\begin{figure}
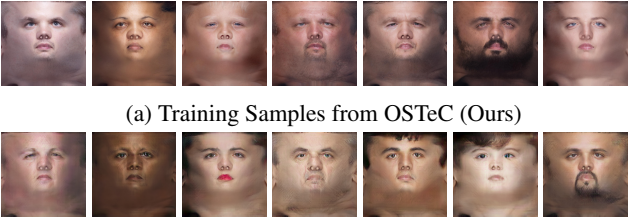

\centering 
\begin{subfigure}{0.49\textwidth}
\foreach \i in {1,2,3,4,5,6,7}{
\includegraphics[width=\var\linewidth]{utem/training/\i}}
\tiny\caption{Training Samples from OSTeC (Ours)}\end{subfigure}
\begin{subfigure}{0.49\textwidth}
\foreach \i in {1,2,3,4,5,6,7}{
\includegraphics[width=\var\linewidth]{utem/generated/\i}}
\tiny\caption{Generated samples from Unsupervised Texture Model (Ours)}\end{subfigure}
\caption{We build a texture model from completed textures from 2D image and train a GAN similar to GANFit~\cite{gecer2019ganfit} approach for high-quality texture modeling.}
\label{fig:utem}
\end{figure}

\subsection{Quantitative Results}

\noindent{\bf UV Texture Completion.}
For the quantitative evaluation of UV texture completion, we employ the UVDB (Multi-PIE \cite{gross2010multi}) dataset released by \cite{deng2018uv}. Following \cite{deng2018uv}, we skip the first 200 subjects, as there is no training, and test on the remaining 137 subjects. We employ two metrics namely peak signal-to-noise ratio (PSNR) and structural similarity index (SSIM), which are computed between the predicted UV texture and the ground truth. In Tab.~\ref{table:UVCompletion7views}, the proposed method shows great priority over UV-GAN \cite{deng2018uv} and CE \cite{pathak2016context}, especially for the profile faces.

\begin{table}
\begin{center}
\resizebox{\linewidth}{!}{
\begin{tabular}{c|c|c|c|c|c}
\hline
Methods & Metric & $0^{\circ}$ & $\pm30^{\circ}$  & $\pm60^{\circ}$ & $\pm90^{\circ}$ \\
\hline
\multirow{2}{*} {CE~\cite{pathak2016context}}  & PSNR & 23.03  & 21.93  & 20.27  & 19.63 \\
                        & SSIM & 0.9201 & 0.8920 & 0.8881 & 0.7179 \\ 
                        \hline
\multirow{2}{*} {UV-GAN \cite{deng2018uv}} & PSNR & 23.36  & 22.25  & 20.53  & 19.83 \\
                         & SSIM & 0.9241 & 0.8971 & 0.8919 & 0.7250 \\
                        \hline
\multirow{2}{*} {Ours}   & PSNR & 23.95 & 22.54  & 21.04  & 20.44  \\
                         & SSIM & 0.9282 & 0.9018 & 0.8979 & 0.7462 \\
\hline
\end{tabular}
}
\end{center}
\vspace{-2mm}
\caption{Quantitative evaluations of UV texture completion on the MultiPIE dataset \cite{gross2010multi} under view changes.}
\label{table:UVCompletion7views}
\vspace{-4mm}
\end{table}

\begin{table}
\small
\begin{center}
\begin{tabular}{c|c|c}
\hline
Method & Frontal-Frontal & Frontal-Profile \\
\hline
Human                                               &96.24 $\pm$ 0.67 & 94.57 $\pm$ 1.10\\
\hline
DR-GAN~\cite{tran2017disentangled}                  & 97.84 $\pm$ 0.79 & 93.41 $\pm$ 1.17\\
DR-GAN+~\cite{tran2018representation}               & 98.36 $\pm$ 0.75 & 93.89 $\pm$ 1.39\\
% Peng \etal~\cite{peng2017reconstruction}            & 98.67            & 93.76           \\
PIM \cite{zhao2018towards}                          & 99.44$\pm$0.36   & 93.10 $\pm$ 1.01\\
% DA-GAN \cite{zhao20183d}                            & 99.48 $\pm$ 0.36 & 95.96$\pm$ 0.85 \\
HF-PIM \cite{cao2019towards}                        & -                & 94.71 $\pm$ 0.83\\
\hline
UVGAN \cite{deng2018uv}                             & 98.83 $\pm$ 0.27 & 93.09  $\pm$ 1.72\\
+Profile2Frontal                                    &-                & 93.55  $\pm$ 1.67\\
+Frontal2Profile                                    &-                & 93.72  $\pm$ 1.59\\
+Set2set                                            &-                & 94.05  $\pm$ 1.73\\
\hline
CASIA-R18-ArcFace                                   & 99.34 $\pm$ 0.49&  93.69 $\pm$ 1.33\\
+Profile2Frontal                                    &-                &  94.87 $\pm$ 0.96\\
+Frontal2Profile                                    &-                &  95.68 $\pm$ 0.91\\
+Set2set                                            &-                &  95.92 $\pm$ 0.87\\
\hline
MS1M-R18-ArcFace                                    & 99.68  $\pm$0.29  &  96.14 $\pm$ 1.06\\
+Profile2Frontal                                    &-                  &  97.06 $\pm$ 0.74\\
+Frontal2Profile                                    &-                  &  97.43 $\pm$ 0.61\\
+Set2set                                            &-                  &  97.85 $\pm$ 0.57\\
\hline
\end{tabular}
\end{center}
\vspace{-2mm}
\caption{Verification accuracy($\%$) comparison on the CFP dataset \cite{sengupta2016frontal}. }
\vspace{-2mm}
\label{table:CFP}
\end{table}

\noindent{\bf Pose-invariant Face Matching.}
We evaluate the performance of frontalization of our work on pose-invariant face recognition in the wild. We choose the widely used dataset CFP \cite{sengupta2016frontal}, which focuses on extreme pose face verification. We employ the ArcFace loss~\cite{deng2019arcface} to train the ResNet-18 networks \cite{zhou2020rotate} on CASIA-WebFace~\cite{yi2014learning} and the refined version of MS1M~\cite{guo2016ms,deng2018arcface}. Note that the backbone of our embedding network is smaller than LightCNN-29 \cite{wu2018light} used by HF-PIM \cite{cao2019towards} and ResNet-27 used by UV-GAN \cite{deng2018uv}.
As shown in Tab.~\ref{table:CFP}, synthesising frontal faces from profile faces significantly improves the accuracy by $1.18\%$ and $0.92\%$ for the ArcFace models trained on CASIA and MS1M, respectively. 
Since face frontalization is a very challenging problem, we also synthesise profile faces from frontal faces following \cite{deng2018uv}, which leads to even better results, $95.68\%$ for the CASIA model and $97.43\%$ for the MS1M model. 
In addition, we use a view interpolation of $15^\circ$ to generate a set of images for each test face. Then, we use the generated set centres to conduct verification. The accuracy further improves to $95.92\%$ for the CASIA model and $97.85\%$ for the MS1M model, both surpassing recent state-of-art methods (\eg HF-PIM \cite{cao2019towards} and UV-GAN \cite{deng2018uv}) by a large margin.  

\subsection{Ablation Study}
We performed an ablation study to explore the contribution of each loss terms in Fig.~\ref{fig:ablation}. The study shows that encoder $\mathcal{E}$ starts with a good initialization. $\mathcal{L}_p$ helps to match some low-level features. $\mathcal{L}_{lan}$ aligns generated images to the input geometry, \eg background leakage around the neck. $\mathcal{L}_{per}$ matches mid-level features and finally $\mathcal{L}_{per}$ shows the biggest contribution by precise identity recovery.

\section{Conclusion}
In this paper, we propose an optimization-based one-shot 3D texture completion and frontalization approach by exploiting pretrained 2D image generation networks. Our approach can generate visually remarkable, accurate and identity-resembling complete texture maps and frontalized faces. The experiments show its superiority over other methods by accuracy and face matching at extreme poses.

\clearpage

{\small
\bibliographystyle{ieee_fullname}
\bibliography{references,egbib}
}

\clearpage
\begin{appendices}
%%%%%%%%% BODY TEXT
\section{Algorithm}
We summarize our method in Algorithm~\ref{alg:algorithm} where the notations are described in the original paper.

\begin{algorithm}[h]
\SetAlgoLined
\SetKwInOut{Input}{Input}
\SetKwInOut{Output}{Output}
\Input{RGB Face Image: $\mathbf{I}_0$}
\Input{3DMM Fitting: $(\mathbf{S}, \mathbf{c})$}
\Input{Novel Camera Views: $\{\mathbf{c}_i\}, i \in \{0 \dots n\}$}
\Output{Completed UV Texture Map: $\overline{\mathbf{T}}_n$}

 $\mathbf{T}_0 \leftarrow \mathcal{R}'(t_{coord}, \mathbf{I}_0, \mathbf{S'})$\\
 \For{$i\gets0$ \KwTo $n$}{
 $\mathbf{S'}_i \leftarrow \mathcal{P}(\mathbf{S}, \mathbf{c}_i)$\\
 $\mathbf{V}_i \leftarrow \text{diag}( \frac{[\mathbf{S}'_i, \mathbf{h}]}{||[\mathbf{S}'_i, \mathbf{h}]||_2} \cdot \mathcal{N}(\mathbf{S}_i)^T)$\\
 }
 \For{$i\gets0$ \KwTo $n$}{
  $\overline{\mathbf{V}}_i \leftarrow   \bigcap\limits_{i \ne j} (\mathbf{V}_i > \mathbf{V}_j)$\\
 $\mathbf{M}^{UV}_i \leftarrow \big((\mathbf{V}_0 > t_1) \cap (2\mathbf{V}_0 > \mathbf{V}_i)\big) \cup \displaystyle\bigcup\limits_{i>j} \overline{\mathbf{V}}_j$\\
 $\mathbf{M}_i \leftarrow  \mathcal{R}(\mathbf{S'}_i, \mathbf{M}^{UV}_i, t_{coord})$\\
 $\mathbf{I}_i \leftarrow \mathcal{R}(\mathbf{S'}_i, \overline{\mathbf{T}}_{i-1}, t_{coord})$\\
 $\mathbf{W}_i \leftarrow \mathcal{E}(\mathbf{I}_i)$\\
 $\mathbf{W}_i \leftarrow  \argmin_{\mathbf{W}_i} \mathcal{L}_{total} (\mathbf{I}_i, \mathbf{M}_i, \mathbf{W}_i) $\\
 $\mathbf{G}_i \leftarrow \mathcal{G}(\mathbf{W}_i)$\\
 $\mathbf{T}_i \leftarrow \mathcal{R}'(t_{coord}, \mathbf{G}_i, \mathbf{S'}_i)$\\
 $\overline{\mathbf{T}}_i \leftarrow \overline{\mathbf{V}}_i \odot \mathbf{T}_i +  (1 - \overline{\mathbf{V}}_i) \odot \overline{\mathbf{T}}_{i-1} $
 }
 \caption{One-Shot Texture Completion}
 \label{alg:algorithm}
\end{algorithm}

\section{Pose-Invariant Face Matching: MultiPIE dataset}

For the evaluation in {\bf under-controlled} scenario, we compare our method with recent state-of-the-art studies, \eg
CPF \cite{yim2015rotating}, DR-GAN \cite{tran2017disentangled}, FF-GAN \cite{yin2017face}, TP-GAN \cite{huang2017beyond}, CAPG-GAN \cite{hu2018pose}, PIM \cite{zhao2018towards}, HF-PIM \cite{cao2018learning} and Rotate $\&$ Render \cite{zhou2020rotate}, on the Multi-PIE dataset \cite{gross2010multi}. 
The performances are reported following the protocol of the setting 2 \cite{yim2015rotating,cao2018learning} provided by the Multi-PIE dataset.
Each testing identity has one gallery image from the first appearance. Hence, there are 72,000 and 137 images in the probe and gallery sets, respectively. 
In Tab. \ref{table:multipieid}, results are reported across different poses.
We employ the strategy of ``recognition via generation''  and faces at any pose are first frontalized by our model.
After the face frontalization, the pre-trained ArcFace model trained on MS1M is employed as the feature extractor. 
Here, we refer to \cite{zhou2020rotate} to train ResNet-18, which is slightly smaller than LightCNN-29 \cite{wu2018light} used by \cite{cao2018learning}.
For those poses less than \({60^{\circ}}\), the performances of most methods are quite good whereas our method almost achieves zero failure rate.
However, the profiles with extreme poses ($> 60^{\circ}$) on are very challenging. For those extreme poses, our method obviously outperforms 
other methods, surpassing the ``Rotate $\&$ Render'' method \cite{zhou2020rotate} by $0.84\%$ under the pose of $90^{\circ}$.
This impressive recognition performance undoubtedly confirms the effectiveness of the proposed identity-preserved UV texture completion.

\begin{table}
\centering
\resizebox{\linewidth}{!}{
\begin{tabular}{ccccccc}
\toprule
Method& \(\pm 15^{\circ}\) & \(\pm 30^{\circ}\) & \(\pm 45^{\circ}\) & \(\pm 60^{\circ}\) & \(\pm 75^{\circ}\) & \(\pm 90^{\circ}\)\\
\midrule
CPF \cite{yim2015rotating} & 95.0 &88.5 &  79.9 & 61.9 &  - & -\\ 
DR-GAN \cite{tran2017disentangled}& 94.9 & 91.1 & 87.2 & 84.6 & -     & -   \\
FF-GAN \cite{yin2017face}& 94.6 & 92.5 & 89.7 & 85.2 & 77.2 & 61.2 \\
TP-GAN \cite{huang2017beyond}& 98.7 & 98.1 & 95.4 & 87.7 & 77.4 & 64.6 \\
CAPG-GAN \cite{hu2018pose} & 99.8 & 99.6 & 97.3 & 90.3 & 83.1 & 66.1 \\
PIM \cite{zhao2018towards} & 99.3 & 99.0 & 98.5 & 98.1 & 95.0 & 86.5 \\
HF-PIM \cite{cao2018learning} & 99.99 & 99.98 & 99.88 & 99.14 & 96.40& 92.32 \\
R$\&$R \cite{zhou2020rotate}  &- &  {\bf 100} &  {\bf 100} & {\bf 99.7} & 99.3 & 94.4\\ 
\midrule
Baseline   & 99.98 &99.86    & 99.80  & 98.50 & 96.19 & 92.06 \\ 
Ours       & {\bf 100}   & {\bf 100}      & 99.88  & 99.62 &  {\bf 99.35} & {\bf 95.24} \\ 
\bottomrule
\end{tabular}
}
\caption{Rank-1 recognition rates (\%) across views on the Multi-PIE dataset \cite{gross2010multi}. The baseline model is ResNet-18 trained on MS1M with the ArcFace loss. Our method further employs face finalization to improve the accuracy.}
\label{table:multipieid}
\end{table}

\section{Performance on `in-the-wild' Scenario}
Following `Pose-Invariant Face Matching' experiment in the original paper, we visualize some of the frontal-profile pairs from CFP dataset~\cite{sengupta2016frontal} to evaluate and verify quantitative experiments qualitatively. Figures~\ref{fig:cfp1},\ref{fig:cfp2},\ref{fig:cfp3} show many pairs of frontal and profile images of the same identity, completed texture UV maps by our method, its rendering, frontalization by our method and cosine similarity scores. The scores are obtained by a ResNet-18 networks~\cite{he2016deep} on CASIA-WebFace~\cite{yi2014learning} for:
\begin{itemize}
    \item \textbf{`Org.':} the pairs of original images
    \item \textbf{`UV.':} original frontal image and rendered geometry with a completed UV map by our method
    \item \textbf{`Frontalized':} original frontal image and frontalized image by our method
\end{itemize}

As can be seen in the figures both qualitatively and quantitatively, our approach can generate excellent quality frontal images and UV texture maps with preserved identity, even under low resolution, extreme pose, occlusion, lighting and expression variations. The cosine similarity scores are mostly improved by the generations of our method compared to the original profile images which verifies the qualitative results.

\def \var {1.0}
\begin{figure*}
\centering
\vspace{-0.2cm}
\begin{subfigure}{0.495\linewidth}
\foreach \i in {004_11-0.2721-0.5243-0.4069,
008_13-0.3327-0.4258-0.5298,
031_13-0.2333-0.3221-0.3979,
038_13-0.3170-0.4840-0.5972,
045_13-0.2814-0.3933-0.4860,
047_12-0.3279-0.7150-0.6226,
058_14-0.2443-0.3230-0.4350,
060_13-0.3408-0.6294-0.6295,
077_12-0.2799-0.4578-0.5110,
093_12-0.2921-0.4374-0.4793}{
\includegraphics[width=\var\linewidth]{suppmat/cfp/\i}\vspace{-0.1cm}
\mymacro{\i}}\end{subfigure}
\begin{subfigure}{0.495\linewidth}
\foreach \i in {095_14-0.2467-0.3714-0.5629,
107_12-0.1910-0.3520-0.4144,
110_11-0.3069-0.4580-0.5054,
111_14-0.2817-0.3825-0.5592,
124_13-0.3257-0.1216-0.3159,
125_13-0.2839-0.3537-0.5617,
134_12-0.3441-0.6188-0.6307,
136_12-0.3025-0.5550-0.4817,
137_12-0.3000-0.3952-0.4683,
207_14-0.2668-0.5800-0.4946}{
\includegraphics[width=\var\linewidth]{suppmat/cfp/\i}\vspace{-0.1cm}
\mymacro{\i}}\end{subfigure}
\caption{Qualitative verification of the \textit{Pose-Invariant Face Matching} experiment. Each block respectively consists of: (1) Original frontal image, (2) Original profile image, (3) Rendered geometry with our completed texture map, (4) Our completed texture map, (5) Our frontalized image.}
\label{fig:cfp1}
\end{figure*}

\def \var {1.0}
\begin{figure*}
\centering
\vspace{-0.2cm}
\begin{subfigure}{0.495\linewidth}
\foreach \i in {212_12-0.3270-0.4017-0.4985,
246_12-0.2314-0.2621-0.3455,
252_11-0.3239-0.5294-0.5108,
276_12-0.3220-0.3749-0.4507,
276_13-0.2815-0.5524-0.5627,
277_12-0.3178-0.2679-0.5406,
282_14-0.3535-0.4687-0.5833,
289_11-0.3611-0.5225-0.4586,
291_13-0.2899-0.5659-0.4974,
300_11-0.2847-0.2742-0.4550}{
\includegraphics[width=\var\linewidth]{suppmat/cfp/\i}\vspace{-0.1cm}
\mymacro{\i}}
\end{subfigure}
\begin{subfigure}{0.495\linewidth}
\foreach \i in {310_14-0.3251-0.1906-0.4698,
349_14-0.2903-0.5230-0.3744,
359_12-0.2347-0.2016-0.4063,
362_11-0.2444-0.0798-0.4252,
386_13-0.3665-0.5696-0.5975,
395_12-0.2750-0.3866-0.4999,
405_14-0.3230-0.3584-0.4714,
413_11-0.3066-0.3104-0.5051,
416_11-0.3558-0.4814-0.5172,
419_13-0.3376-0.5806-0.5417}{
\includegraphics[width=\var\linewidth]{suppmat/cfp/\i}\vspace{-0.1cm}
\mymacro{\i}}
\end{subfigure}
\caption{Qualitative verification of the \textit{Pose-Invariant Face Matching} experiment. Each block respectively consists of: (1) Original frontal image, (2) Original profile image, (3) Rendered geometry with our completed texture map, (4) Our completed texture map, (5) Our frontalized image.}
\label{fig:cfp2}
\end{figure*}

\def \var {1.0}
\begin{figure*}
\centering
\vspace{-0.2cm}
\begin{subfigure}{0.495\linewidth}
\foreach \i in {438_11-0.2931-0.3546-0.4556,
441_13-0.3472-0.6477-0.5133,
442_12-0.3028-0.4160-0.4935,
448_13-0.3061-0.5749-0.6450,
452_14-0.2548-0.3515-0.4516,
462_12-0.2417-0.2932-0.4577,
474_11-0.3625-0.5053-0.6189,
490_14-0.3291-0.3974-0.5685,
499_13-0.2689-0.4482-0.5187,
343_11-0.3400-0.3351-0.3944}{
\includegraphics[width=\var\linewidth]{suppmat/cfp/\i}\vspace{-0.1cm}
\mymacro{\i}}
\caption{Successful cases}
\label{fig:cfp3}
\end{subfigure}
\begin{subfigure}{0.495\linewidth}
\foreach \i in {005_12-0.2858-0.1868-0.3251,
121_11-0.3269-0.1960-0.3145,
130_13-0.3146-0.0173-0.3831,
145_13-0.3221-0.8032-0.6299,
200_11-0.2807-0.2053-0.3055,
220_11-0.2542-0.3466-0.3391,
224_13-0.3827-0.3984-0.4523,
231_12-0.3083-0.1267-0.3100,
299_11-0.3155-0.2631-0.3359,
323_14-0.2483-0.1801-0.3249}{
\includegraphics[width=\var\linewidth]{suppmat/cfp/\i}\vspace{-0.1cm}
\mymacro{\i}}
\caption{Failure cases}
\label{fig:failure}
\end{subfigure}
\vspace{-0.2cm}
\caption{Qualitative verification of the \textit{Pose-Invariant Face Matching} experiment. Each block respectively consists of: (1) Original frontal image, (2) Original profile image, (3) Rendered geometry with our completed texture map, (4) Our completed texture map, (5) Our frontalized image.}
\end{figure*}

\section{Manipulating Frontalized Faces}
Frontalization by our approach is achieved by rendering the geometry that is textured by the completed UV map and reconstructing it in StyleGAN~\cite{karras2018style} latent space. Therefore the frontalized images can be manipulated by common StyleGAN manipulation techniques such as interpolation between different identities and changing/adding some facial attributes. 

Fig.~\ref{fig:slow_frontalization} illustrates some interpolations performed between the original and the frontalized projections that slowly shift from various poses to the frontal pose. Fig.~\ref{fig:interpolation} shows interpolation between different identities, both in the frontalized and the original projections. Please note that, the frontalized interpolation maintain smoother transition between the identities, whereas the original image projections generates artefacts at the intermediate generations due to exhausted latent parameters. Lastly, Fig.~\ref{fig:manipulation} illustrates attribute manipulation by extracting some attribute directions with~\cite{shen2019interpreting} such as age, gender and expression.

\def \var {1.0}
\begin{figure*}
\centering
\foreach \i in {teaser_3,
Amber_Tamblyn_0001,
Benjamin_Netanyahu_0005,
im2,
im18,
Norah_Jones_0003,
Robin_Wright_Penn_0001,
teaser_2,
teaser_4,
teaser_5,
teaser_6,
teaser_7}{
\includegraphics[width=\var\linewidth]{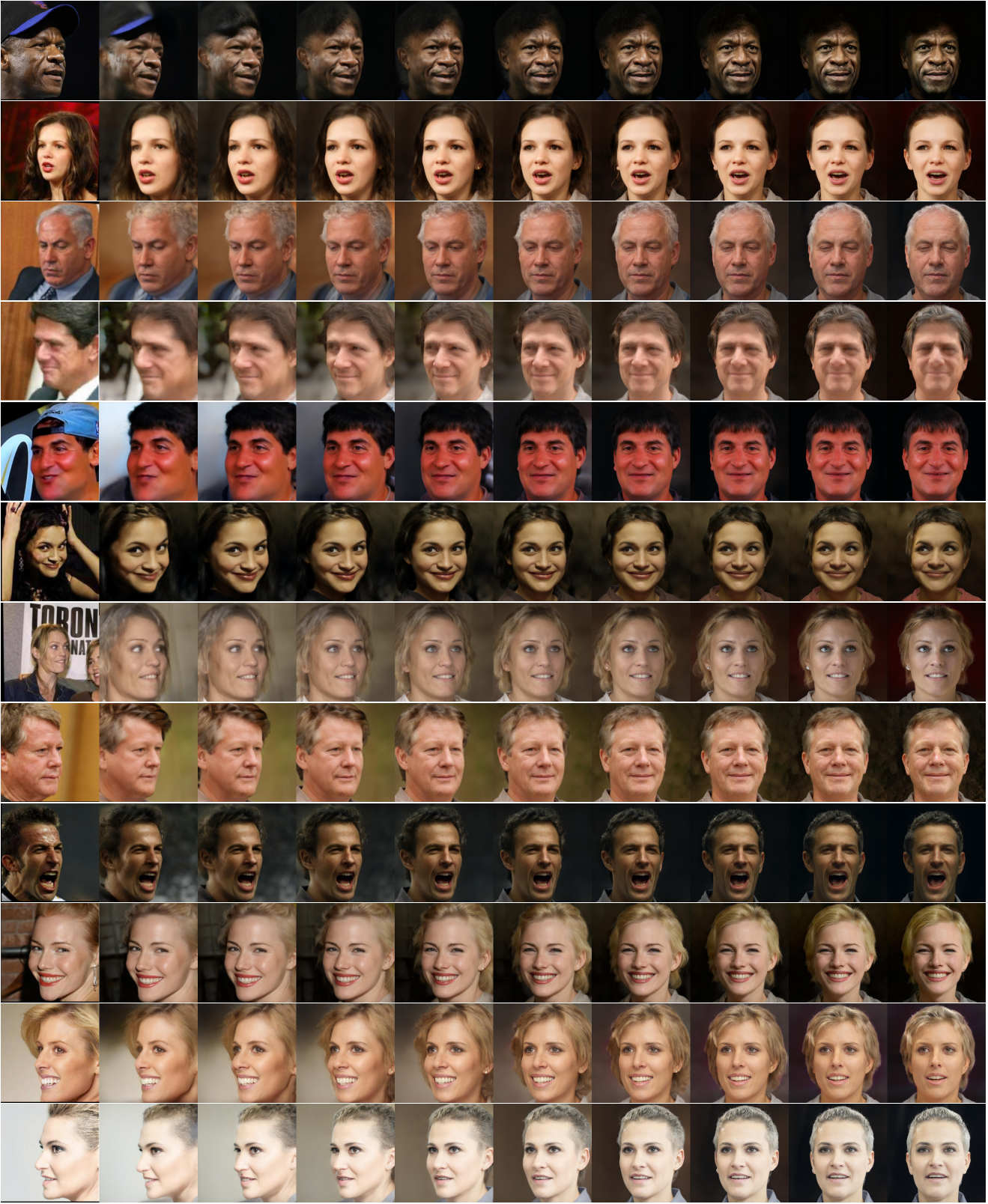}}
\caption{Interpolation between the original poses and the frontalized versions. First column is the original image. Second column is its projection to the StyleGAN space. Last column is the frontalized generated image by our approach. And other columns are the interpolation in-between.}
\label{fig:slow_frontalization}
\end{figure*}

\def \var {1.0}
\begin{figure*}
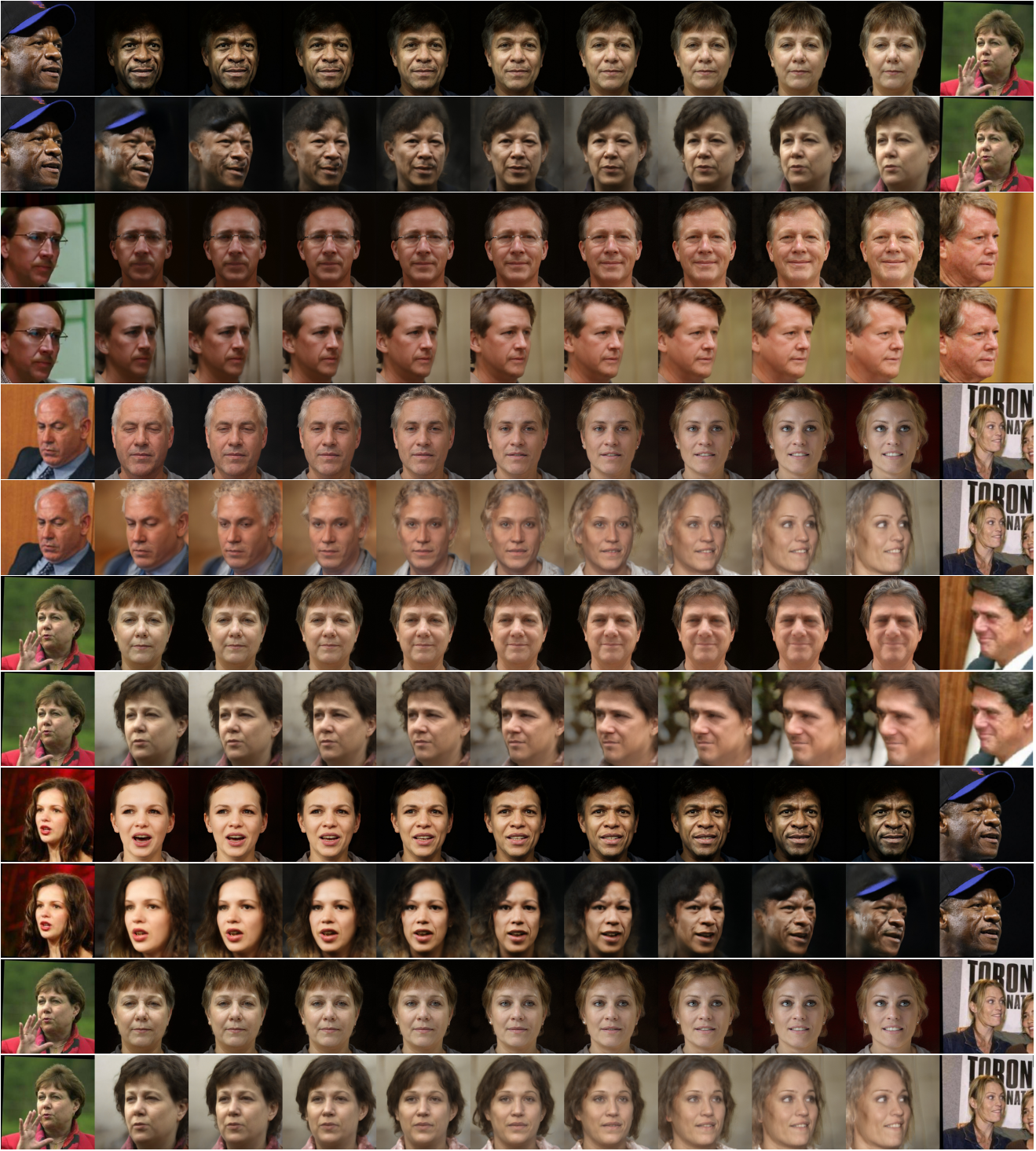

\centering
\foreach \i in {teaser_3_Ann_Veneman_0010,
im16_teaser_2,
Benjamin_Netanyahu_0005_Robin_Wright_Penn_0001,
Ann_Veneman_0010_im2,
Amber_Tamblyn_0001_teaser_3,
Ann_Veneman_0010_Robin_Wright_Penn_0001}{
\includegraphics[width=\var\linewidth]{suppmat/interpolation/\i}
\includegraphics[width=\var\linewidth]{suppmat/interpolation_org/\i}}
\caption{Interpolations between different identities. First and Last colums are the original images and other columns are interpolations. Odd rows are interpolating frontal projections and even rows are interpolating the original image projections. Please note that, the frontalized interpolation maintain smoother transition between the identities, whereas the original image projections generates artefacts at the intermediate generations due to exhausted latent parameters.}
\label{fig:interpolation}
\end{figure*}

\def \var {1.0}
\begin{figure*}
\centering
\foreach \i in {age,gender,smile}{
\includegraphics[width=\var\linewidth]{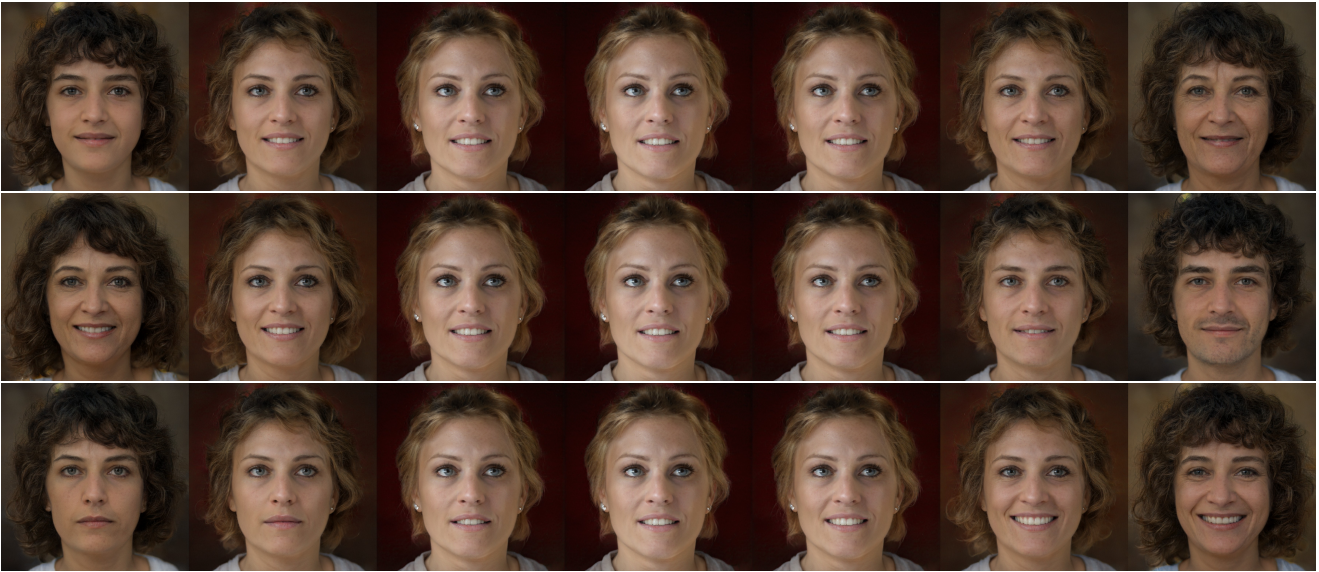}}
\caption{Attribute manipulation by~\cite{shen2019interpreting} can be performed on the frontalized images.}
\label{fig:manipulation}
\end{figure*}

\section{Limitations and Failure Cases}
The biggest strength and the biggest weakness of our approach is being an optimization-based method. Usually, the running time takes around 5-10 minutes depending on the convergence speed. This is mainly due to CPU intensive visibility mask and 3D mesh rendering over the iterations. We believe that the code might be optimized to run under 1 minute which is a reasonable running time for an optimization-based method.

Another limitation of an optimization-based method is the danger of local minima. We observed in some cases, optimization gets stuck at local minima, failing to find a good texture completion and frontalization. This is partially addressed by the encoder network $\mathcal{E}$, but emprically we can still observe this behaviour as can be seen in Fig.~\ref{fig:failure}.

Another drawback of our approach is that it heavily relies on 3D face reconstruction. Therefore, our method is limited by the accuracy and performance of the 3D reconstruction. That is to say, if some part of the identity cannot be captured by the reconstruction, our method might struggle to compensate. Some of such failure cases are illustrated in Fig.~\ref{fig:failure}.

\end{appendices}

\end{document}